%% file: acl_latex.tex
\pdfoutput=1

\documentclass[11pt]{article}

\usepackage[preprint]{acl}
\usepackage{times}
\usepackage{latexsym}

\usepackage[T1]{fontenc}

\usepackage[utf8]{inputenc}

\usepackage{microtype}

\usepackage{inconsolata}

\usepackage{amsfonts}  
\usepackage{subfigure}
\usepackage{verbatim}
\usepackage{enumitem}
\usepackage{tikz}
\usepackage{multirow}
\usepackage{pifont}
\usepackage{amssymb}
\usepackage{bm}
\usepackage{amsmath}
\usepackage{booktabs}
\usepackage{makecell}
\usepackage{colortbl}
\usepackage{arydshln}
\usepackage{array}
\usepackage{graphicx}
\usepackage[figuresright]{rotating}
\usepackage{xspace} 
\usepackage{bbold}
\usepackage{tabularx}
\usepackage[normalem]{ulem}
\usepackage{diagbox}
\usepackage{amsmath}

\definecolor{pinegreen}{RGB}{15,153,15}
\definecolor{mygray}{gray}{.9}
\definecolor{target}{RGB}{0,0,146}
\definecolor{myblue}{RGB}{25,101,255}
\definecolor{myorange}{RGB}{239,134,63}

\newcommand*{\circled}[1]{\lower.7ex\hbox{\tikz\draw (0pt, 0pt)%
		circle (.5em) node {\makebox[1em][c]{\small #1}};}}
\newcommand{\tabincell}[2]{\begin{tabular}{@{}#1@{}}#2\end{tabular}}

\newcommand{\myroman}[1]{\uppercase\expandafter{\romannumeral#1}}

\makeatletter
\newcommand*\bigcdot{\mathpalette\bigcdot@{.5}}
\newcommand*\bigcdot@[2]{\mathbin{\vcenter{\hbox{\scalebox{#2}{$\m@th#1\bullet$}}}}}
\makeatother

\newcommand{\coke}{\textbb{COKE}\xspace}
\newcommand{\cokelm}{\textbb{COLM}\xspace}
\newcommand{\cokewithemoji}{\textbb{COKE}\xspace}

\newcommand{\emojisituation}{\includegraphics[height=1em,trim=0 3em 0 0]{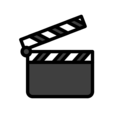}\xspace} 
\newcommand{\emojiclue}{\includegraphics[height=1em,trim=0 3em 0 0]{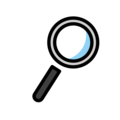}\xspace} 
\newcommand{\emojithought}{\includegraphics[height=1em,trim=0 3em 0 0]{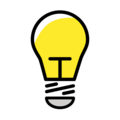}\xspace}
\newcommand{\emojiemotion}{\includegraphics[height=1em,trim=0 3em 0 0]{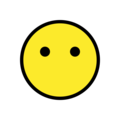}\xspace}
\newcommand{\emojiaction}{\includegraphics[height=1em,trim=0 3em 0 0]{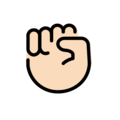}\xspace}

\usepackage[T1]{fontenc}    
\usepackage{url}            
\usepackage{booktabs}       
\usepackage{nicefrac}       
\usepackage{microtype}      
\usepackage{xcolor}         

\definecolor{empathy}{RGB}{168,201,234}
\definecolor{elicitation}{RGB}{245,203,177}
\definecolor{negative}{RGB}{199,231,171}
\definecolor{positive}{RGB}{255,212,196}

\definecolor{situationcolor}{RGB}{242,242,242}
\definecolor{cluecolor}{RGB}{248,203,173}
\definecolor{thoughtcolor}{RGB}{255,242,204}
\definecolor{actioncolor}{RGB}{222,235,247}
\definecolor{emotioncolor}{RGB}{226,240,217}

\definecolor{cluecolor1}{RGB}{253, 208, 178}
\definecolor{cluecolor2}{RGB}{255, 213, 183}
\definecolor{cluecolor3}{RGB}{255, 218, 188}

\title{{\cokewithemoji}: A Cognitive Knowledge Graph for Machine Theory of Mind}

\author{Jincenzi Wu$^{1,2}$\thanks{\quad Work done during internship at the CoAI Group.}\thanks{\quad Equal Contribution.} \quad Zhuang Chen$^1$\footnotemark[2]  \quad Jiawen Deng$^3$ \quad Sahand Sabour$^1$\\
\textbf{Helen Meng$^2$} \quad \textbf{Minlie Huang$^1$}\thanks{\quad Corresponding author.}\\
$^1$CoAI Group, DCST, IAI, BNRIST, Tsinghua University \\
$^2$The Chinese University of Hong Kong, Hong Kong SAR, China \\ 
$^3$University of Electronic Science and Technology of China \\
{jincenziwu@gmail.com \quad zhchen-nlp@mail.tsinghua.edu.cn \quad aihuang@tsinghua.edu.cn}}

\begin{document}
\maketitle
\begin{abstract}
Theory of mind (ToM) refers to humans' ability to understand and infer the desires, beliefs, and intentions of others. 
The acquisition of ToM plays a key role in humans' social cognition and interpersonal relations.
Though indispensable for social intelligence, ToM is still lacking for modern AI and NLP systems since they cannot access the human mental state and cognitive process beneath the training corpus. To empower AI systems with the ToM ability and narrow the gap between them and humans, in this paper, we propose \cokewithemoji: the first cognitive knowledge graph for machine theory of mind. Specifically, \coke formalizes ToM as a collection of 45k+ manually verified cognitive chains that characterize {human mental activities} and subsequent behavioral/affective responses when facing specific {social circumstances}. 
In addition, we further generalize \coke using LLMs and build a powerful generation model \cokelm tailored for cognitive reasoning. Experimental results in both automatic and human evaluation demonstrate the high quality of \coke, the superior ToM ability of \cokelm , and its potential to significantly enhance social applications. We release our code and data at \url{https://github.com/jincenziwu/COKE}.


\end{abstract}

\section{Introduction}

\begin{figure}[t]
	\centering
	\vspace{0mm}
        \hspace{0mm}
	\setlength{\abovecaptionskip}{0mm}
	\setlength{\belowcaptionskip}{0mm}
        \includegraphics[width=0.50\textwidth]{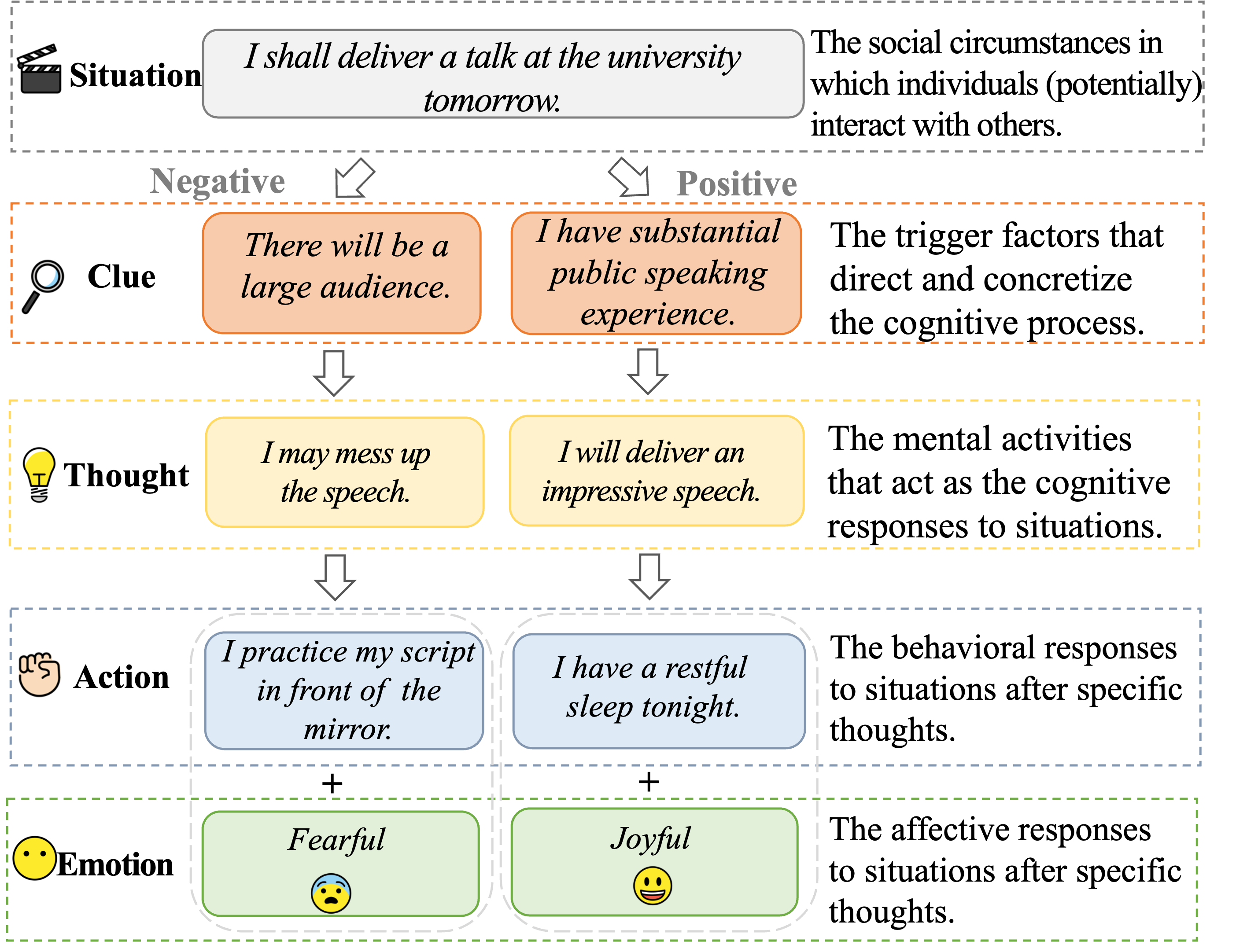}
	\centering
	\caption{\coke instantiates Theory of Mind as positive and negative cognitive chains in social situations. \textit{ Situation{$\Rightarrow$} Clue {$\Rightarrow$} Thought{$\Rightarrow$} (Action $+$ Emotion).} }

	\label{fig-intro}
	\vspace{-4mm}
\end{figure}

In social environments, human beings must be able not only to react to what others are doing, but also to anticipate what they will do. This ability to understand and infer human goals is typically described as Theory of Mind (ToM) \cite{premack1978tom}.
One way of accomplishing ToM is to observe what others do in various situations, and derive a set of affective and behavioral rules.
When the same or highly similar things arise again, we can bring out plausible predictions accordingly \cite{call2011tom30year}.
Figure \ref{fig-intro} presents an example that \emph{someone will deliver a talk at the university tomorrow} (a social circumstance), and \emph{he has substantial public speaking experience} (a trigger factor). We can plausibly anticipate that \emph{he will deliver an impressive speech} (a mental activity), \emph{feels joyful} (an affective response), and \emph{has a restful sleep tonight} (a behavioral response). Here ToM is instantiated as a chained cognitive process that derives from our knowledge, experiences, and memories \cite{harris1989young}.
ToM is indispensable to humans since it allows us to leverage our own minds to simulate others', so as to achieve efficient communication \cite{RabinowitzPSZEB18}.

Despite its importance for social intelligence, ToM is not well internalized by modern AI and NLP systems. \citet{shapira2023clever} illustrates a significant decline in performance and outright failure of Large Language Models (LLMs) in ToM tasks, particularly evident when confronted with adversarial samples. 
The main reason is that learning-based systems are usually trained on superficial text corpora, while lacking access to the underlying human mental state and cognitive process \cite{Sap2022NeuralTO}.  
In other words, NLP systems rely on the maximum likelihood to understand and generate texts, but do not go beneath the surface to the desires, beliefs, and intentions of humans.

In this paper, we introduce \cokewithemoji : the first \emph{\textbf{CO}gnitive \textbf{K}nowledg\textbf{E} graph} for machine theory of mind. Our goal is to formalize ToM and make it accessible and learnable for AI systems. In \coke, we instantiate ToM as a collection of manually verified cognitive chains that characterize humans' mental activities in specific social circumstances along with their behavioral and affective responses \cite{meinhardt2018two,mehl2020theory}. 
Each cognitive chain involves five types of nodes: 
\textbf{1) \textit{situations}} denote the social circumstances;
\textbf{2) \textit{clues}} denote the trigger factors;
\textbf{3) \textit{thoughts}} denote the mental activities;
\textbf{4) \textit{actions}} denote the behavioral responses;
\textbf{5) \textit{emotions}} denote the affective responses.
Moreover, as shown in Figure \ref{fig-intro}, individuals react differently to the same situation due to the diversified cognitive processes. 
Therefore, for each situation, we derive multiple cognitive chains and further label them as \textbf{\textit{positive}} (means optimistic) or \textbf{\textit{negative}} (means pessimistic) to mark the chain polarity.
We propose to induce the raw data from LLMs, and then recruit educated workers majoring in psychology for manual selection and revision. The resulting knowledge graph constitutes 62,328 nodes and 45,369 cognitive chains.

The construction of \coke offers the basic ToM ability to understand and infer the human goals in already collected situations \cite{call2011tom30year}. But obviously, it is impossible to enumerate all situations in the real world. Thus we move one step further and build a cognitive language model \cokelm to cope with unseen situations that have not appeared in the knowledge graph. Specifically, we decompose the construction of cognitive chains into four cognitive generation tasks, then finetune LLMs using the manually collected data in \coke. By this means, we combine the commonsense knowledge embedded in LLMs and the ToM ability provided by \coke, enabling \cokelm to infer cognitive chains for unseen situations. 

We summarize our contributions in this work as follows. \textbf{1)} We propose the first cognitive knowledge graph for machine theory of mind. We instantiate human theory of mind as a collection of 45k$+$ manually verified cognitive chains, which provides a basic ToM ability for accessing and learning. \textbf{2)} We build a powerful cognitive language model \cokelm by associating \coke with LLaMA-2 \cite{touvron2023llama}, so as to predict cognitive chains for out-of-KG situations. \textbf{3)} We conduct extensive experiments to evaluate the ToM ability of \cokelm and typical LLMs. The results show that \cokelm outperforms strong baseline models such as GPT-4 \cite{achiam2023gpt} in both zero-shot and few-shot settings, proved by automatic and human evaluations in all cognitive generation tasks, which in turn demonstrates the high quality of \coke. \textbf{4)} We further substantiate the potential of \coke in enhancing social applications, and prove its effectiveness on downstream emotional support conversation task.


\section{\coke: Cognitive Knowledge Graph}

\subsection{Preliminaries}
\label{sec-tom}
\paragraph{What is Theory of Mind?} Theory of Mind refers to the ability of humans to understand and infer other people's desires, beliefs, and intentions \cite{premack1978tom}. 
Under specific social circumstances (\textbf{\textit{situations}}), the core mechanism by which ToM empowers us is to ascribe others' {mental activities} (\textbf{\textit{thoughts}}), and predict their corresponding {behavioral responses} (\textbf{\textit{actions}}) and {affective responses} (\textbf{\textit{emotions}}) \cite{leslie2004core,apperly2010mindreaders}. Furthermore, the acquisition of ToM enables us to realize and appreciate that people can have different cognitive responses in the same situation due to various {trigger factors} (\textbf{\textit{clues}}) like personality and experience \cite{meinhardt2018two}.

\paragraph{Why Do AI Systems Need ToM?} 

ToM has been a persistent yet elusive goal of artificial intelligence for decades \cite{choi2022curious}. AI systems need ToM to understand a user's situations and predict subsequent reactions, so as to provide effective responses or operations that meet the user's needs \cite{Le2019RevisitingTE, Dhelim2021IoTEnabledSR, Langley2022TheoryOM}. However, recent studies \cite{shapira2023clever} show that today's LLMs still lack ToM and social intelligence. The main reason is that AI systems can only learn from text corpora in training, but cannot access the human mental state and cognitive process that determine what and why to say. Now, there are neither public resources that contain all the concepts in ToM, nor commonsense knowledge graphs that depict the structure of cognitive chains in ToM. This motivates us to propose the first cognitive knowledge graph for machine Theory of Mind.

\begin{figure*}[h]
	\centering
	\vspace{0mm}
	\hspace{-10mm}
	\setlength{\abovecaptionskip}{1mm}
	\setlength{\belowcaptionskip}{1mm}
	\includegraphics[width=0.90\linewidth]{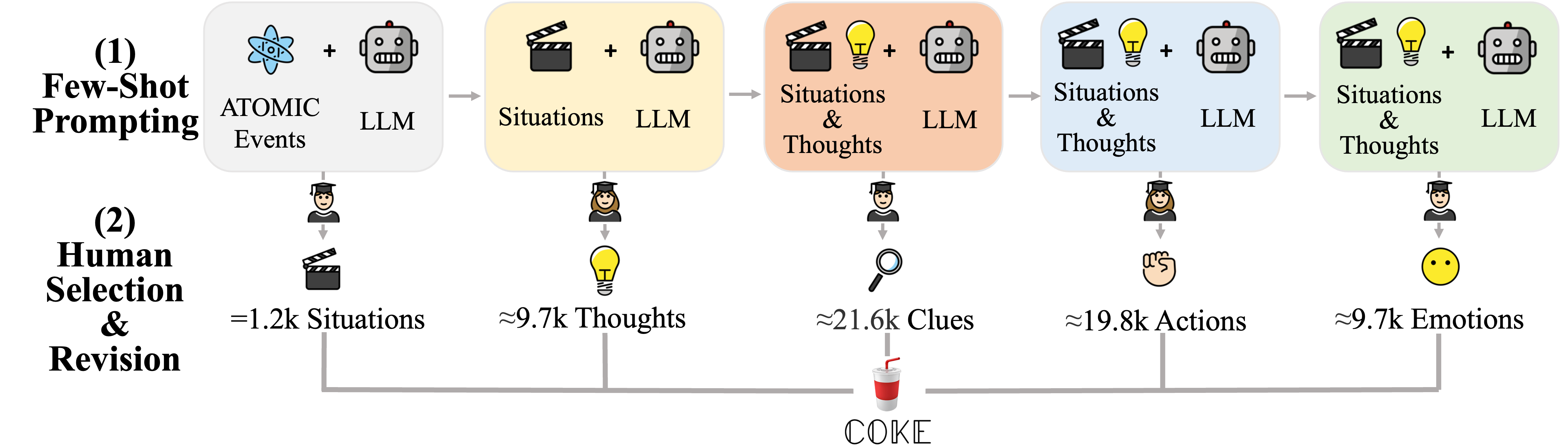}
	\centering
	\caption{The two-step data collection approach for constructing \coke.}
	\label{fig-datacollection}
	\vspace{-4mm}
\end{figure*}

\subsection{Data Structure of \coke}

According to the above-mentioned psychology research on theory of mind, we specify five types of nodes in \coke: \textit{\textbf{situations}}, \textit{\textbf{clues}}, \textit{\textbf{thoughts}}, \textit{\textbf{actions}}, and \textit{\textbf{emotions}}. 
We here define the basic unit of \coke as the following cognitive chain: \textit{Situation{$\Rightarrow$}Clue{$\Rightarrow$}Thought{$\Rightarrow$}(Action$+$Emotion)}. This structure depicts the intact cognitive process of ToM: When a person faces a situation, some clues trigger his/her thoughts, along with his/her actions and emotions. Furthermore, to distinguish whether a cognitive chain is optimistic or pessimistic under the specific situation, we further define its polarity as \textbf{\textit{positive}} or \textbf{\textit{negative}}. In practice, the polarity of the cognitive chain is determined by its thought node. Notice that, in \coke, we omit the definition of edges (i.e., the connections between nodes) since they can be easily inferred when the types of nodes are already known. 
We then illustrate the nodes in detail.

\textbf{{\emojisituation} Situations} in \coke denote the social circumstances in which individuals (potentially) interact with others. By referring to DailyDialog \cite{LiSSLCN17}, a widely used daily social dialogue dataset, we select the five most common social topics: \textit{School} (what happened at school), \textit{Work} (what happened at work), \textit{Tourism} (travel and entertainment), \textit{Relationship} (social activities between individuals), \textit{Ordinary Life} (what happened in families). Detailed information about each topic can be found in Appendix \ref{appendix-topicinfo}.



\textbf{{\emojiclue}Clues} in \coke denote the trigger factors that direct and concretize the cognitive process. In a specific situation, humans' mental activities are triggered and directed by relevant subjective and objective factors \cite{meinhardt2018two}. According to the taxonomy from \cite{baldwin1992relational}, clues mainly involve the particular information about personality, knowledge, experience, education, objective facts, social rules, and so on.

 \textbf{{\emojithought}Thoughts} in \coke denote the mental activities that act as the cognitive responses to situations. Thoughts serve as the bridge between the external environment and individual cognition, thus can be considered as the core of ToM \cite{Westbrook2007AnIT}. 
As mentioned before, the polarity of a cognitive chain is anchored to its thought node. In other words, an optimistic thought marks the entire cognitive chain as positive, and a pessimistic thought marks it as negative.

 \textbf{{\emojiaction}Actions} in \coke denote the behavioral responses to situations after specific thoughts. Notice that the semantic meaning of actions may not conform to their polarity annotations. For example, in Figure \ref{fig-intro}, 
 the action ``\textit{I practice my script in front of the mirror}'' is a neutral sentence. However, it is the consequence of the negative thought ``\textit{I may mess up the speech}'', so it is still labeled as negative.


\textbf{{\emojiemotion}Emotions} in \coke denote the affective responses to situations after specific thoughts. Without loss of generality, we restrict the emotions to six basic categories \cite{shaver1987emotion}: \textit{Love}, \textit{Surprise}, \textit{Joyful}, \textit{Sad}, \textit{Angry}, and \textit{Fearful}. The first three appear in positive cognitive chains, while the last three appear in negative cognitive chains.



\subsection{Data Creation and Selection}
Recent studies have proved that LLMs trained on huge text corpora can naturally serve as a repository for data collection \cite{west2023novacomet,kim2023soda}. 
Inspired by their successful attempts, we propose a two-step data collection approach for constructing \coke. As shown in Figure \ref{fig-datacollection}, \textbf{1)} we first manually design suitable few-shot prompts to induce \texttt{GPT-3.5} (i.e., text-davinci-002 ) \citet{ouyang2022training} to automatically generate raw data for five types of nodes in a pipeline manner. \textbf{2)} We then recruit and train eight graduate students majoring in social psychology as annotators to select and revise the outputs of \texttt{GPT-3.5}. Next, we illustrate the details of how to prompt \texttt{GPT-3.5}, then introduce the data statistics after human annotation. 
More detailed parameters and templates used for prompting are provided in Appendix \ref{appendix-gpt3para} and \ref{appendix-gpt3prompt}.

\paragraph{Prompting LLMs for {\emojisituation}Situations}
{{Situations}} in \coke denote the social circumstances in which individuals (potentially) interact with others, which exhibit a strong correlation with social events and describe the social environment of daily lives. Unfortunately, to the best of our knowledge, there is no public dataset that contains a wide variety of social situations. Therefore, we choose the events in ATOMIC \cite{Sap2019ATOMICAA}, a frequently-used commonsense knowledge graph for if-then reasoning, as an alternative. However, the events from ATOMIC are still not qualified to become the situations in \coke. The reasons are two-fold, and we here take the event ``\textit{PersonX misses calls from PersonY}'' as an example. \textbf{1)} ATOMIC events replace the characters with ``\textit{PersonX}'' and ``\textit{PersonY}'', thereby losing most of the interpersonal information that is indispensable for social situations. For example, if someone misses calls from his employer, he may worry that something is wrong with his work. But if the caller changes to his girlfriend, he may sense love because she cares about him. \textbf{2)} ATOMIC events omit most of the social context information, making the background environments where the events occur not available. For example, we don't know if the calls in the above event happen at work or on vacation, so we have no idea how to reify subsequent cognitive processes.

\begin{figure}[h]
	\centering
	\vspace{-1mm}
	\hspace{0mm}
	\setlength{\abovecaptionskip}{1mm}
	\setlength{\belowcaptionskip}{1mm}
	\includegraphics[width=0.48\textwidth]{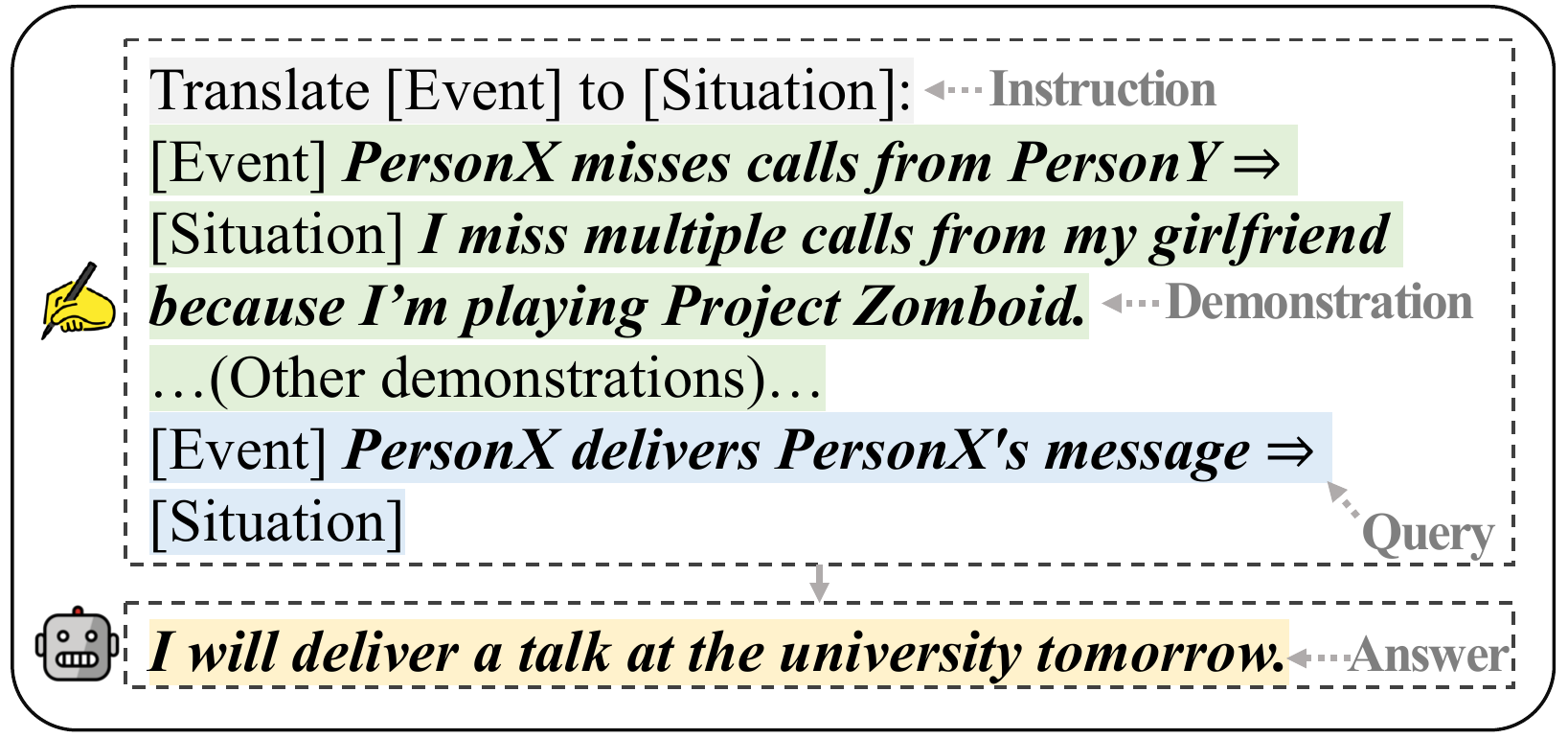}
	\centering
	\caption{Prompting LLMs for situations.}
	\label{fig-data-situation}
	\vspace{-2mm}
\end{figure}

To address the problem, we choose to prompt \texttt{GPT-3.5} to rewrite ATOMIC events to qualified \coke situations. Specifically, we first manually create several demonstration (input, output) pairs such as (``\textit{PersonX misses calls from PersonY}'', ``\textit{I miss multiple calls from my girlfriend because I’m playing Project Zomboid}''). Then we wrap the demonstrations with the task instruction (``\textit{Translate [Event] to [Situation]}'') and an arbitrary event query (``\textit{PersonX delivers PersonX's message}'') to form an input template as shown in Figure \ref{fig-data-situation}. 
As a result, we collect the output situation from \texttt{GPT-3.5} like ``\textit{I will deliver a talk at the university tomorrow}''. 
In practice, we manually select 400 ATOMIC events that are general and easy to adapt, then rewrite each event to 5 situations with different topics. Finally, we obtain \textbf{2,000} raw situations.

\paragraph{Prompting LLMs for {\emojithought}Thoughts} Thoughts in \coke denote the mental activities that act as the cognitive responses to situations. As mentioned above, thoughts serve as the bridge between the external environment and individual cognition, and can be seen as the core of ToM. Since thoughts are triggered by certain clues, they come after clues in cognition chains. However, our pilot experiments show that directly using situations to prompt LLMs can better stimulate their generation ability and get more diverse thoughts, which is beneficial to manual selection and revision. Therefore, here we temporarily reverse the order of thoughts and clues, i.e., we first prompt LLMs using situations to generate thoughts, and then prompt LLMs using thoughts to generate clues.


	




Similar to the instruction prompts for collecting situations, we manually construct the demonstration as ``\textit{When [Situation], I feel great/terrible since I think [Thought]}'' and again wrap it with the task instruction and an arbitrary situation query to form the input template. Notice that we use different sentiment terms (``\textit{great}'' and ``\textit{terrible}'') to control the polarity (\textit{positive} and \textit{negative}) of the generated thoughts. Since the polarity of a cognitive chain is anchored to its thought, we have also controlled the polarity of the entire cognitive chain that will be generated. 
After prompting, we obtain \textbf{14,400} raw thoughts, half of which are positive and the other half are negative.

\paragraph{{Prompting LLMs for {\emojiclue}Clues, {\emojiaction}Actions, and \emojiemotion}Emotions}
Clues in \coke denote the trigger factors that direct and concretize the cognitive process. Thus we construct the demonstrations as `` \textit{Complete the sentence: When [Situation], I think [Thought] since [Clue]}'' to prompt LLMs for clues. Actions and emotions in \coke denote the behavioral and affective responses to situations after specific thoughts. We hence construct the demonstrations as ``\textit{When [Situation], I think [Thought] so [Action]}'' and ``\textit{When [Situation], I think [Thought] and I feel [Emotion]}'' to prompt LLMs for actions and emotions. Since emotion belongs to predefined categories, we further modify its corresponding prompt to the question-answering form. Finally, we obtain \textbf{29,364} raw clues, \textbf{29,364} raw actions, and \textbf{14,400} raw emotions. Their polarities are already determined by previously generated thoughts.

\paragraph{Human Selection and Revision}
After collecting the raw data, we manually annotate a small amount of data and formulate several rules to distinguish good data from bad data. Subsequently, we use the detailed definition of nodes in \coke and the filtering rules as a tutorial to train the eight annotators. After passing an annotation qualification test, they are asked to select and revise the raw data of five types of nodes. 
 More details about human annotation can be found in Appendix \ref{appendix-humanannotation}. 

As shown in Table \ref{tab-statistics}, the final reserved data contains \textbf{1,200} situations, \textbf{9,788} thoughts, \textbf{21,677} clues, \textbf{19,875} actions, and \textbf{9,788} emotions, resulting in an overall retention rate of around 70\%. This statistic proves that ToM ability exhibited by most powerful LLMs like \texttt{GPT-3.5} is still not satisfactory enough even with delicate prompting, thus further emphasizing the necessity of our construction of \coke. After linking and ordering the obtained nodes, we instantiate ToM with a total of \textbf{45,369} cognitive chains in \coke, containing \textbf{23,252} positive chains and \textbf{22,117} negative chains in English. Example \coke data can be found in Appendix \ref{appendix-dataexample}.
\input{tables/data_statistic.tex}

\input{tables/cokelm_flow.tex}






\section{Cognitive Language Model \cokelm}

By consulting the constructed cognitive knowledge graph \coke, we can obtain the basic ability of theory of mind (ToM) via matching the faced situation to a similar situation in KG, and then inspecting the involved cognitive chains (a.k.a, entity linking). But obviously, \coke only collects finite situations and cannot cover the infinite and diverse situations in the real world. Inspired by the methods for automatic knowledge graph completion like COMET \cite{BosselutRSMCC19}, we propose a cognitive language model \cokelm to cope with unseen situations and expand the scope of application of \coke. \cokelm is built upon LLMs, aiming to integrate the commonsense knowledge inside LLMs and the ToM ability from \coke. To this end, we first decompose the cognitive process into a sequence of cognitive generation tasks, and then finetune LLaMA-2 \cite{touvron2023llama} using the collected data from \coke. 

\vspace{-2mm}
\subsection{Cognitive Generation Tasks}
In \coke, a cognitive process towards theory of mind (ToM) is instantiated as a sequenced cognitive chain containing situation, clue, thought, action, and emotion. Therefore, given a situation, we can decompose the cognitive process into four generation tasks as shown in Table \ref{tab-cokelm}. These four tasks work in a pipeline manner, and the complete cognitive chain can be restored by linking their generated results.
1) \textbf{Clue Generation}. 
When facing a specific situation, humans can automatically distinguish the factors that may influence beliefs and trigger thoughts. 
2) \textbf{Thought Generation}. 
In a specific situation, the related clues trigger and arouse diversified human mental activities, i.e., thoughts. 
3) \textbf{Action Generation}.
Driven by specific thoughts, we humans will take corresponding actions to realize our beliefs and achieve our goals. 
Notice that we omit clues here since their impacts are largely covered by the triggered thoughts. 4) \textbf{Emotion Generation}.
After forming a specific thought in a certain situation, humans will naturally generate corresponding emotions to express attitudes and views on the situation. 
Since emotions are limited to 6 categories, this task is a classification task.

By linking the above four tasks in a pipeline manner, we can restore the cognitive chain ``\textit{Situation {$\Rightarrow$} Clue {$\Rightarrow$} Thought {$\Rightarrow$} (Action$+$Emotion)}'', thus preserving the complete cognition process of ToM. Since each cognitive chain in \coke is labeled with polarity, each cognitive generation task can be further divided into positive and negative subtasks (e.g., positive thought generation and negative thought generation).


\subsection{Training \cokelm}
After decomposing the cognitive chain into four cognitive generation tasks, we can process the data in \coke accordingly to obtain training samples. For computational efficiency, we design \cokelm as a multi-task controllable generation model, so that it can simultaneously accomplish four cognitive generative tasks and further control the polarity (i.e., positive or negative) of the outputs. As shown in Table \ref{tab-cokelm}, for each task and polarity, we insert specific tokens (i.e.,[NegClue],[PosClue]) in the input $X$ to guide the generation process of \cokelm. For implementation, \cokelm is built as a decoder-only architecture and initialized with the LLaMA-2 \cite{touvron2023llama}. 
The model is trained with each input-output pair $X$-$Y$ from any task.

\section{Experiments}
%
In this section, we construct a dataset from \coke to evaluate ToM ability of our cognitive language model \cokelm, and compare it with advanced LLMs including GPT-3.5 Turbo, GPT-4, LLaMA-2-7B (our backbone) and Mistral-7B. We first illustrate the experimental setup, then analyze the experimental results for automatic and human evaluation, and finally validate \coke's effectiveness to empower social applications.


	

\subsection{Experimental Setup}



To evaluate the ToM ability of different models, we randomly split 1,200 social situations into 1,080 (90\%) training and 120 (10\%) validation situations. 
Then we can automatically split samples for different cognitive generation tasks according to the situations, and obtain training/validation splits as 19,409/2,268 (clue), 8,746/1,042 (thought), 17,982/1,893 (action), and 8,746/1,042 (emotion). Based on this setting, we ensure that the cognitive chains in the validation set all occur in \textbf{UNSEEN} situations, which is crucial for ToM evaluation. 

For \cokelm, we use the Hugging Face implementation\footnote{{https://huggingface.co/meta-llama/Llama-2-7b-hf}.} of LLaMA-2, and train it for 20 epochs using the AdamW optimizer \cite{LoshchilovH19} and LoRA \cite{hu2021lora} with learning rate 3e-4 and batch size 32 in Five Tesla V100 GPUs. The whole training process costs about 9 GPU hours. For baseline models, we construct manual prompts to enable them to complete the cognitive generation tasks. We evaluate 0-shot, 2-shot, and 5-shot performance of all baseline LLMs. More details can be found in Appendix \ref{experiment-prompt}. Due to the data deficiency, we report the performance on the validation set. 


\input{tables/automatic_evaluation4}

\subsection{Main Results}

\paragraph{Automatic Evaluation} To evaluate the clue/thought/action generation tasks, we use METEOR \cite{LavieD09}, ROUGE \cite{lin2004rouge}, BLEU-1, BLEU-2 \cite{Papineni2002BleuAM} and BERTScore \cite{zhang2019bertscore} as metrics. In these three tasks, each input may be mapped to multiple ground truth outputs. 
Therefore, following \citet{mostafazadeh-etal-2020-glucose}, we compute the average scores between all predicted sequences and all ground truth outputs\footnote{Here we provide all models with the required polarity (positive or negative) via special tokens or prompts.}. Moreover, to evaluate the emotion generation (classification) task, we compute the classification accuracy as the metric. We present all automatic evaluation results in Table \ref{tab-autoeval}.
Compared to baseline models like LLaMA-2 and Mistral, which face challenges in following task instructions, \cokelm shows significant performance enhancements across various cognitive generation tasks with \coke. While leveraging additional prompts boosts performance for powerful models like GPT-3.5 Turbo and GPT-4 via in-context learning in clue/thought/action generation tasks, these LLMs still struggle with complex emotional understanding \cite{wang2023emotional} and fail in the emotion classification task. Compared to these powerful LLMs, \cokelm maintains substantial advantages across all evaluation metrics. We hereby have two observations: 1) The data we collected in \coke is of high quality and can empower the model with strong ToM ability. 2) \cokelm, which is designed as a controllable generative model for multiple cognitive tasks, can effectively internalize the ToM ability and cope with unseen situations.


\paragraph{Human Evaluation} Beyond automatic evaluation, we also wonder how humans perceive the cognitive chains generated by different models since ToM is essentially a human ability. Therefore, for each model, we sample a cognitive chain for each situation in the validation set (resulting in 90 samples per model) for human evaluation. We present three experts with the cognitive chains generated by \cokelm and top-performing LLMs in automatic evaluation, and then ask them to score on a 5-point Likert scale based on \textit{relevance}, \textit{correctness}, and \textit{logicality}. Specifically, they were instructed to rate the chains as follows: ``completely irrelevant (1 point)'', ``contains irrelevant content (2 points)'', ``relevant but does not follow the task definition correctly (3 points)'', ``follows the task definition correctly but does not logically continue the cognitive chain (4 points)'' and ``continues the cognitive chain correctly and logically (5 points)''. We calculate the final score for each model by averaging these ratings. A rating of 5 indicates \textbf{valid} content. 

The results in Table \ref{tab-hueval} illustrate that the cognitive chains generated by \cokelm are more acceptable to humans, which further demonstrates the superiority of its ToM ability. 
As a further illustration, we present ten exemplary cognitive chains generated by \cokelm across five diverse situations, covering various topics, in Appendix \ref{appendix-casestudy}.

\input{tables/Human_evaluation}

\subsection{Empowering Social Application} 
As \coke is a cognitive knowledge graph for machine theory of mind, we further substantiate its effectiveness in empowering social applications with ToM capabilities. 
Our chosen testing ground is the emotional support conversation (ESC) task \cite{DBLP:conf/acl/LiuZDSLYJH20}. The prime objective of ESC is to generate empathetic and effective responses in social dialogues, with the aim of mitigating users' emotional distress and fostering improved mental states. ESC requires ToM because it involves understanding and responding to a user's emotional state and underlying challenges by understanding their mental processes and intentions.



\paragraph{Integration of \coke in ESC}

The ESC dataset has already provided the {situation} behind each dialogue. We then treat each user's utterance in the dialogue as a {clue}, and employ \cokelm to infer two thoughts (1 positive and 1 negative) via thought generation. 
Based on the generated thoughts, we can accordingly infer
actions for each thought via action generation.
We simply extract verbs and nouns in thoughts or actions as the ToM knowledge keywords, and append them to the end of dialogue history, serving as an enhanced context. Based on the context, the ESC system is trained to generate the corresponding response. We here use the Blenderbot \cite{DBLP:conf/eacl/RollerDGJWLXOSB21} as the generation model, which is the same as the original ESC paper. 


\paragraph{ESC Performance with ToM Knowledge}

In our experiments, we compare three models: Vanilla ESC, ESC with \cokelm Actions, and ESC with \cokelm Thoughts. We conduct both automatic and human evaluations, following the same methodology as the original ESC paper. 
We sample 30 dialogues in the validation set and let two experts rate \textit{Fluency}, \textit{Identification}, \textit{Comfort}, and \textit{Suggestions} on a 3-point Likert scale. The final score is the average of these ratings.
The experimental results shown in Table \ref{tab-esc-autoevaluation} demonstrate that, when incorporating the ToM knowledge into the response generation process, both the ESC with \cokelm Actions and ESC with \cokelm Thoughts models significantly outperform the Vanilla ESC model across all metrics. For clarity, we further present a case study and how \coke offers a more nuanced and empathetic approach to AI-driven emotional support in Appendix \ref{appendix-esc} .

\input{tables/AppendixH_ESC_AE}

\section{Related Work}
\vspace{-0.5mm}
Cognitive knowledge is essential for modeling intuitive tasks, such as reasoning, in language models. Previous work mainly focuses on realizing cognitive knowledge as an accessory to commonsense knowledge. They construct text-based knowledge graphs that include numerous instances of Head-Relation-Tail triplets regarding different events and objects.
One of the widely-used examples of such knowledge graphs is ConceptNet \cite{Speer2016ConceptNet5A}, which provides a graph of concepts connected by relations, covering a variety of taxonomic facts. \citet{Sap2019ATOMICAA} proposed ATOMIC, a graph of if-then inferences that models social commonsense in daily life events. \citet{hwang2021symbolic} expanded upon ATOMIC by incorporating two additional categories of commonsense relations: physical-entity commonsense relations and event-centered commonsense relations. \citet{west2021symbolic} prompted GPT-3 to distill commonsense knowledge and create a machine-generated corpus ATOMIC\textsuperscript{10x}. \citet{kim2023soda} further leveraged it to generate a large-scale dataset focused on socially-grounded conversations. Furthermore, NOVATOMIC \cite{west2023novacomet} employs natural language queries as open relations to link the commonsense knowledge, enabling its application in general reasoning tasks.




Generally, existing KGs consider event-centered commonsense relations, such as temporal and causal relationships, represented as \(Event\overset{\text{IsBefore}}{\xleftarrow{\hspace*{1cm}}} Event\overset{\text{Causes}}{\xrightarrow{\hspace*{1cm}}}Event\).
However, they have not explicitly addressed ToM concepts and relations, which are crucial for accessing and interpreting human mental states and cognitive processes. In contrast, \coke delineates ToM concepts and structures them as a cognitive chain: \textit{ Situation{$\Rightarrow$} Clue{$\Rightarrow$} Thought{$\Rightarrow$}(Action$+$ Emotion)}. This chained structure is designed to mirror human cognitive processes, enabling a deeper understanding of how individuals infer others' mental states in specific social circumstances along with their behavioral and affective responses. Moreover, the ATOMIC family, which uses short phrases with generic placeholders like ``PersonX'', significantly constrains the social and interpersonal information critical for complex ToM reasoning \cite{zalla2018prior}.  \coke addresses this limitation by providing a richer and more nuanced context for each cognitive concept in the chain. This approach broadens the scope of cognitive knowledge accessible to machine systems and enhances their capability in social cognition, hereby fostering downstream applications.

\section{Conclusion}
\vspace{1.5mm}
In this work, we present \cokewithemoji, the first cognitive knowledge graph for machine theory of mind. We instantiate ToM as a series of cognitive chains to describe human mental activities and behavioral/affective responses in social situations.
Through prompting \texttt{GPT-3.5} and manual annotation, we collect 62k+ nodes and construct 45k+ cognitive chains.
Based on \coke, we build a powerful cognitive language model \cokelm. \cokelm can handle unseen situations and predict complete cognitive chains in a pipeline manner. Automatic and human evaluations show that \cokelm effectively internalizes the ToM ability and outperforms strong baseline models like GPT-4. 

We further demonstrate that \coke can empower LLMs with a nuanced understanding of human affective and cognitive reasoning, resulting in superior performance in ESC. By integrating \coke into LLMs, they gain the ability to comprehend and predict human cognitive processes with greater depth. We believe that \coke and \cokelm can facilitate social applications such as dialogue systems and autonomous agents.



\section*{Limitations}
In this work, we introduce \cokewithemoji, the first cognitive knowledge graph for machine Theory of Mind, which aims to empower AI systems with cognitive capabilities. However, we acknowledge the following omissions and inadequacies in our work.
\paragraph{Data Coverage}
We acknowledge that there is a limitation in topic coverage of proposed \cokewithemoji, which only covers five social topics that are commonly discussed.
Besides, \cokewithemoji has a relatively small data scale in comparison with other relative knowledge graphs, due to its more specific node contents and more complicated construction process.
Consequently, \cokewithemoji cannot cover all situations in deployment, and the cognitive reasoning models constructed on this basis may have unreliable predictions in out-of-domain situations.
\paragraph{Cognitive Inference Ability}
As a pioneer in integrating \emph{Theory of Mind} into AI systems, the proposed cognitive language model \cokelm still has a lot of room for improvement in inferring cognitive chains when deployed in practice.
In \cokelm, we take LLaMA-2 as the backbone model to validate the gain of \cokewithemoji on the cognitive inference ability of language models.
We acknowledge that adopting larger backbone models would contribute to a more powerful inference model, which is worth exploring in more depth in the future.

\paragraph{}

\section*{Ethics Statement}
We present the first cognitive knowledge graph \cokewithemoji for machine theory of mind.
Our graph is built based on public datasets and model generations.
We strictly adhere to data source usage protocols and ensure that the proposed \cokewithemoji can be released and used \MakeUppercase{legally}.

The construction of \cokewithemoji has gone through the steps of manual selection and revision.
We manually filtered the data containing potentially private information, such as phone numbers and email addresses, to protect user \MakeUppercase{privacy} further. 
We also carefully delete abusive, offensive, biased, and other inappropriate content to avoid unpredictable \MakeUppercase{ethical} hazards.

Our knowledge graph is designed to empower AI systems with the ability of cognitive inference.
We are aware that this capability could be misused in malicious scenarios in the future. However, we believe the value of this work outweighs the risks, and we also call for more socially responsible research in this field.

\bibliography{coke}
\bibstyle{acl}

\clearpage
\appendix
\input{sections/08-appendix.tex}

\end{document}

%% file: tables/data_statistic.tex


\begin{table}[h]
\vspace{0mm}
\small
\renewcommand\arraystretch{1.05}
\setlength{\abovecaptionskip}{1mm}
\setlength{\belowcaptionskip}{1mm}
\centering
\setlength{\tabcolsep}{1.0mm}

\begin{tabular}{@{}lrrcc@{}}
\toprule    Dimension
          & Raw & Final & \multicolumn{1}{c}{{Avg. Len.}} & Retention Rate   \\ \midrule
{\emojisituation} Situation & 2,000   & 1,200  &        11.5        &  60.00\%              \\
{\emojithought} Thought   &  14,400  & 9,788 &        6.6         &  67.97\%           \\
{\emojiclue} Clue      & 29,364  & 21,677&        7.3         &  73.82\%           \\
{\emojiaction} Action    & 29,364  & 19,875 &        6.8         &  67.68\%           \\
{\emojiemotion} Emotion   & 14,400  & 9,788 &        1.0           &  67.97\%           \\ \bottomrule
\end{tabular}
\caption{The statistics of \coke.}
\label{tab-statistics}
\vspace{-2mm}
\end{table}

%% file: tables/cokelm_flow.tex
\begin{table*}[h]
\vspace{0mm}
\small
\renewcommand\arraystretch{1.0}
\setlength{\abovecaptionskip}{0mm}
\setlength{\belowcaptionskip}{5mm}
\centering
\setlength{\tabcolsep}{2mm}
\resizebox{0.98\textwidth}{!}{
\begin{tabular}{r|l|l|c}
\hline
\multicolumn{1}{c|}{Task}    & \multicolumn{1}{c|}{Generation Setting}    & \multicolumn{1}{c|}{Input $X$}   & Output $Y$ \\ \hline
Clue Generation    
& \includegraphics[height=1em,trim=0 3em 0 0]{figures/openmoji-situation}Situation$\Rightarrow$\includegraphics[height=1em,trim=0 3em 0 0]{figures/openmoji-clue}Clue   
& Situation \texttt{[NegClue]}                      
& Clue         
\\
Thought Generation 
& \includegraphics[height=1em,trim=0 3em 0 0]{figures/openmoji-situation}Situation + \includegraphics[height=1em,trim=0 3em 0 0]{figures/openmoji-clue}Clue $\Rightarrow$ \includegraphics[height=1em,trim=0 3em 0 0]{figures/openmoji-thought}Thought
& Situation \texttt{[NegClue]}  Clue \texttt{[NegThought]}       & Thought
\\
Action Generation 
& \includegraphics[height=1em,trim=0 3em 0 0]{figures/openmoji-situation}Situation + \includegraphics[height=1em,trim=0 3em 0 0]{figures/openmoji-thought}Thought $\Rightarrow$ \includegraphics[height=1em,trim=0 3em 0 0]{figures/openmoji-action}Action  
& Situation \texttt{[NegThought]} Thought \texttt{[NegAction]}   & Action      
\\
Emotion Generation 
& \includegraphics[height=1em,trim=0 3em 0 0]{figures/openmoji-situation}Situation + \includegraphics[height=1em,trim=0 3em 0 0]{figures/openmoji-thought}Thought $\Rightarrow$ \includegraphics[height=1em,trim=0 3em 0 0]{figures/openmoji-emotion-nomouse}Emotion 
& Situation \texttt{[NegThought]} Thought \texttt{[NegEmotion]} 
& Emotion      \\ \hline
\end{tabular}
}

\vspace{1mm}
\caption{Decomposition of the cognitive process to four cognitive generation tasks for training \cokelm. The complete outputs of four tasks can be restored to cognitive chains. Here \texttt{[Neg.*]} denotes special tokens for controllable generation in \textbf{\textit{negative}} cognitive chains. When coming to \textbf{\textit{positive}} cognitive chains, we use \texttt{[Pos.*]}.
}

\label{tab-cokelm}
\vspace{-9mm}
\end{table*}

%% file: tables/automatic_evaluation4.tex
\begin{table*}[t]
\small
\vspace{0mm}

\centering
\setlength{\abovecaptionskip}{0mm}
\setlength{\belowcaptionskip}{0mm}
\setlength{\tabcolsep}{1.mm}
\resizebox{1.\textwidth}{!}{




\begin{tabular}{llccc|ccc|ccc|ccc|c}
\hline
\multicolumn{1}{c}{\multirow{2}{*}{\textbf{Task}}}                                  & \multicolumn{1}{c}{\multirow{2}{*}{\textbf{Model}}} & \multicolumn{3}{c|}{\textbf{GPT-3.5 Turbo}}          & \multicolumn{3}{c|}{\textbf{GPT-4}}    & \multicolumn{3}{c|}{\textbf{LLaMa-7B}} & \multicolumn{3}{c|}{\textbf{Mistral-7B}} & \textbf{\cokelm} \\
\multicolumn{1}{c}{}                                                                & \multicolumn{1}{c}{}                                & 0-shot               & 2-shot & 5-shot               & 0-shot & 2-shot & 5-shot               & 0-shot      & 2-shot      & 5-shot     & 0-shot       & 2-shot      & 5-shot      &                   \\ \hline
\multirow{5}{*}{\textbf{\begin{tabular}[c]{@{}l@{}}Clue   \\ Gen.\end{tabular}}}  & METEOR↑                   & 0.215                & 0.268  & {\underline{0.293}} & 0.211  & 0.286  & 0.292                & 0.251       & 0.247       & 0.246      & 0.233        & 0.238       & 0.224       & \colorbox{cluecolor}{\textbf{0.370}}    \\
                                                                                    & ROUGE↑                   & 0.222                & 0.245  & 0.266                & 0.190  & 0.260  & {\underline{0.270}} & 0.183       & 0.181       & 0.177      & 0.172        & 0.178       & 0.184       & \colorbox{cluecolor}{\textbf{0.381}}    \\
                                                                                    & BLUE-1↑                   & 0.200                & 0.227  & 0.259                & 0.183  & 0.249  & {\underline{0.269}} & 0.115       & 0.117       & 0.119      & 0.117        & 0.116       & 0.117       & \colorbox{cluecolor}{\textbf{0.340}}    \\
                                                                                    & BLEU-2↑                   & 0.007                & 0.007  & 0.008                & 0.005  & 0.008  & {\underline{0.009}} & 0.003       & 0.003       & 0.003      & 0.003        & 0.003       & 0.004       & \colorbox{cluecolor}{\textbf{0.011}}    \\
                                                                                    & BertScore↑               & 0.862                & 0.868  & 0.872                & 0.859  & 0.871  & {\underline{0.872}} & 0.840       & 0.842       & 0.844      & 0.829        & 0.832       & 0.841       & \colorbox{cluecolor}{\textbf{0.876}}    \\ \hline
\multirow{5}{*}{\textbf{\begin{tabular}[c]{@{}l@{}}Thought   \\ Gen.\end{tabular}}} & METEOR↑                   & 0.166                & 0.266  & {\underline{0.274}} & 0.131  & 0.244  & 0.250                & 0.144       & 0.200       & 0.207      & 0.142        & 0.184       & 0.190       & \colorbox{thoughtcolor}{\textbf{0.305}}    \\
                                                                                    & ROUGE↑                    & 0.132                & 0.205  & {\underline{0.215}} & 0.109  & 0.198  & 0.203                & 0.094       & 0.132       & 0.141      & 0.152        & 0.203       & 0.144       & \colorbox{thoughtcolor}{\textbf{0.344}}    \\
                                                                                    & BLUE-1↑                   & 0.195                & 0.259  & 0.313                & 0.196  & 0.321  & {\underline{0.326}} & 0.121       & 0.115       & 0.112      & 0.117        & 0.164       & 0.160       & \colorbox{thoughtcolor}{\textbf{0.371}}    \\
                                                                                    & BLEU-2↑                   & 0.017                & 0.024  & 0.029                & 0.018  & 0.030  & {\underline{0.031}} & 0.011       & 0.010       & 0.009      & 0.010        & 0.015       & 0.015       & \colorbox{thoughtcolor}{\textbf{0.037}}    \\
                                                                                    & BertScore↑               & 0.874                & 0.890  & {\underline{0.896}} & 0.871  & 0.894  & 0.896                & 0.835       & 0.847       & 0.855      & 0.833        & 0.840       & 0.854       & \colorbox{thoughtcolor}{\textbf{0.902}}    \\ \hline
\multirow{5}{*}{\textbf{\begin{tabular}[c]{@{}l@{}}Action   \\ Gen.\end{tabular}}}  & METEOR↑                   & 0.206                & 0.253  & {\underline{0.291}} & 0.209  & 0.248  & 0.247                & 0.134       & 0.207       & 0.217      & 0.168        & 0.202       & 0.210       & \colorbox{actioncolor}{\textbf{0.342}}    \\
                                                                                    & ROUGE↑                    & 0.186                & 0.255  & {\underline{0.284}} & 0.193  & 0.243  & 0.257                & 0.104       & 0.172       & 0.178      & 0.133        & 0.171       & 0.182       & \colorbox{actioncolor}{\textbf{0.378}}    \\
                                                                                    & BLUE-1↑                   & 0.153                & 0.271  & {\underline{0.302}} & 0.168  & 0.223  & 0.257                & 0.121       & 0.130       & 0.149      & 0.121        & 0.122       & 0.130       & \colorbox{actioncolor}{\textbf{0.378}}    \\
                                                                                    & BLEU-2↑                   & 0.010                & 0.018  & {\underline{0.020}} & 0.011  & 0.014  & 0.017                & 0.008       & 0.008       & 0.009      & 0.007        & 0.008       & 0.008       & \colorbox{actioncolor}{\textbf{0.027}}    \\
                                                                                    & BertScore↑               & 0.872                & 0.882  & {\underline{0.886}} & 0.866  & 0.879  & 0.882                & 0.821       & 0.851       & 0.842      & 0.832        & 0.841       & 0.844       & \colorbox{actioncolor}{\textbf{0.892}}    \\ \hline
\textbf{\begin{tabular}[c]{@{}l@{}}Emotion   \\ Class.\end{tabular}}                & Accuracy↑                 & {\underline{0.693}} & 0.437  & 0.474                & 0.634  & 0.613  & 0.649                & Failed      & 0.068       & 0.062      & Failed       & Failed      & Failed      & \colorbox{emotioncolor}{\textbf{0.793}}    \\ \hline
\end{tabular}

}



\vspace{1mm}
\caption{Automatic evaluation results for cognitive generation tasks. The highest scores are highlighted in color, and second best results are underlined. All results are average scores of 3 runs with random seeds. 
}


\label{tab-autoeval}
\vspace{-4mm}
\end{table*}

%% file: tables/Human_evaluation.tex
\begin{table}[htbp]
\small
\centering
\vspace{0mm}

\setlength{\tabcolsep}{2mm}
\scalebox{1.0}{


\begin{tabular}{lllll}
\hline
\textbf{Task}                                                                      & \textbf{Model}            & \textbf{\begin{tabular}[c]{@{}l@{}}GPT-3.5\\  Turbo\end{tabular}} & \textbf{GPT-4} & \textbf{\cokelm} \\ \hline
\multirow{2}{*}{\textbf{\begin{tabular}[c]{@{}l@{}}Clue   \\ Gen.\end{tabular}}}   & Overall & 4.22   & 4.72           & \colorbox{cluecolor}{4.83}              \\
& Valid.  & 61\%   & 85\%           & \colorbox{cluecolor}{90\%}             \\ \hline
\multirow{2}{*}{\textbf{\begin{tabular}[c]{@{}l@{}}Thought\\ Gen.\end{tabular}}}   & Overall & 4.84   & 4.86           & \colorbox{thoughtcolor}{4.87}              \\
& Valid.  & 82\%   & 88\%           & \colorbox{thoughtcolor}{90\%}              \\ \hline
\multirow{2}{*}{\textbf{\begin{tabular}[c]{@{}l@{}}Action   \\ Gen.\end{tabular}}} & Overall & 4.70   & 4.79           & \colorbox{actioncolor}{4.82}              \\
& Valid.  & 70\%   & 78\%           & \colorbox{actioncolor}{85\%}              \\ \hline
\end{tabular}

}

\vspace{-3mm}
\caption{ Human evaluation results of \cokelm and baseline LLMs with 5 shots. \textbf{Overall}, determining if the response is relevant, correct and logical to continue the cognitive chains. \textbf{Valid.} refers to the validation percentage of outputs.} 

\label{tab-hueval}
\vspace{-4mm}
\end{table}

%% file: tables/AppendixH_ESC_AE.tex

\begin{table}[htbp]
\small
\centering
\vspace{0mm}

\setlength{\tabcolsep}{2mm}
\scalebox{1.0}{


\begin{tabular}{llll}
\hline
Metric          & Vanilla & + Action & + Thought \\ \hline
PPL ↓           & 16.96   & 15.96    & \colorbox{thoughtcolor}{15.88}     \\
BLEU-2 ↑        & 6.93    & 7.18     & \colorbox{thoughtcolor}{7.34}      \\
ROUGE-L ↑       & 15.01   & 15.73    & \colorbox{thoughtcolor}{15.89}     \\
Extrema ↑       & 50.28   & 50.22    & \colorbox{thoughtcolor}{50.43}     \\ \hline
Fluency ↑       & 2.30    & 2.45     & \colorbox{thoughtcolor}{2.60}      \\
Identification↑ & 1.95    & 1.95     & \colorbox{thoughtcolor}{2.30}      \\
Comforting ↑    & 2.10    & 2.35     & \colorbox{thoughtcolor}{2.50}     \\
Suggestion ↑    & 1.35    & 1.40     & \colorbox{thoughtcolor}{2.05}      \\ \hline
\end{tabular}

}

\vspace{-3mm}
\caption{Automatic and human evaluation on ESC.}
\label{tab-esc-autoevaluation}
\vspace{0mm}
\end{table}

%% file: sections/08-appendix.tex

\section{Topic Keywords}
\label{appendix-topicinfo}
We identify the top 8 keywords in five topics using TF-IDF. 
As indicated in Table \ref{tab-keywords}, the information provided by situations in each topic is distinct and relevant to the theme. According to the keywords ``\textit{boss}'' or ``\textit{work}'', we can locate information such as ``\textit{I attend my boss's meeting}'' in the Work topic. The project deadline dilemma like ``\textit{I bet my friend that I can finish my project before the deadline}'' is shared in the School topic. Besides, we can find the travel schedule like ``\textit{I asked my best friend for advice on what to pack for my upcoming trip to Europe}'' in the Tourism topic. Date sharing like ``\textit{I brought my date back to my place after the movie}'' can be found in the Relationship topic. Daily diary like ``\textit{My friend is helping me plan my surprise party}'' appears in the Ordinary Life topic.

\input{tables/AppendixA_keywords.tex}

\section{ Parameters for GPT-3.5 API Utilization}
\label{appendix-gpt3para}
During the data collection process, we use the GPT-3.5 API offered by OpenAI. We read the terms of service\footnote{\href{https://openai.com/api/policies/service-terms/}{https://openai.com/api/policies/service-terms/}.} and follow the usage policies\footnote{\href{https://beta.openai.com/docs/usage-policies}{https://beta.openai.com/docs/usage-policies}.} .
We give the parameter details of the GPT-3.5 API utilized in data collection and the experiments in Table \ref{tab-IntructGPT_parameter}. Our data collection was finalized before the release of GPT-3.5 Turbo and GPT-4. Therefore, we opted for the powerful model \textit{text-davinci-002}.

\input{tables/AppendixD_GPT3.tex}

\section{Prompts for Data Collection}
\label{appendix-gpt3prompt}
In Table \ref{tab-promptforsituation}, we present the detailed prompts for how to extend the base event to the situation in our \coke. When we design the prompt template, we find that the base events prefixed with the token \textbf{[Sentence] }in the prompt lead to better generation than those with the token \textbf{[Event]}. 
Table \ref{tab-promptforgentask} and Table \ref{tab-promptforemotask} show the specific prompts used to collect data in the negative and positive chains for each generation task.



\section{Instruction for Human Selection and Revision }
\label{appendix-humanannotation}

Since our cognitive knowledge graph is closely related to psychology, we choose and train eight graduate students majoring in social psychology as annotators. They are evenly distributed by gender (four males and four females) and come from various regions. 
We pay the annotators approximately \$12 per hour, which exceeds the local minimum wage, and the total cost for annotation is about \$6,000. We illustrate the relevant background and how the data would be used clearly in the first beginning. 

To construct the knowledge graph for machine Theory of Mind (ToM), we have researched relevant papers and books and discussed it with the social psychology professor several times over three months. As a result, we decompose the ToM inference into four inference tasks and clearly define the five types of nodes. In the annotation instructions, explicit definitions of nodes are provided.

\begin{figure*}[h]
	\centering
	\vspace{0mm}
	\hspace{0mm}
	\setlength{\abovecaptionskip}{1mm}
	\setlength{\belowcaptionskip}{1mm}
	\includegraphics[width=0.99\linewidth]{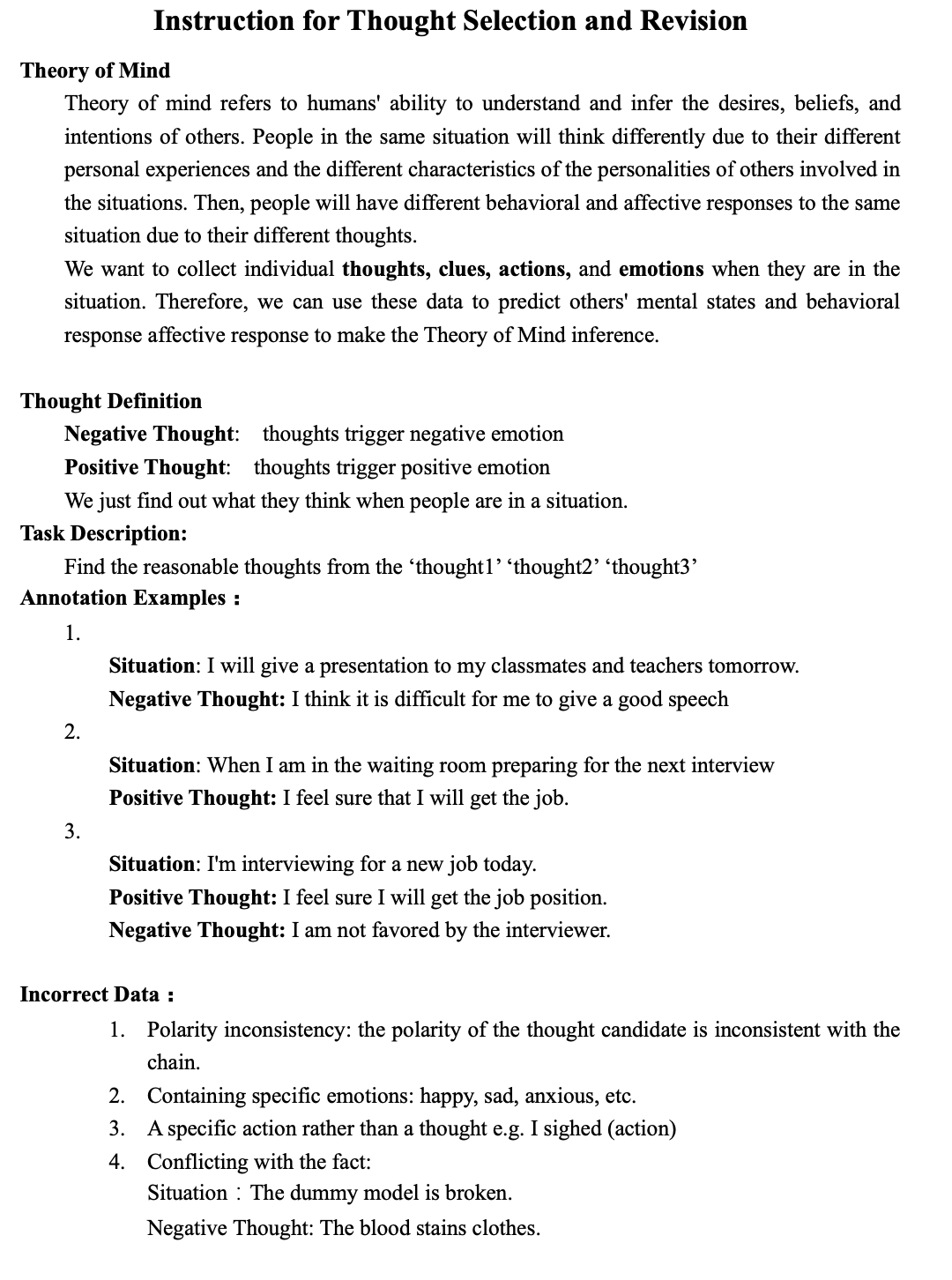}
	\centering
	\caption{Instruction for Thought selection and revision. }
	\label{fig-thought_instruction}
\end{figure*}

\begin{figure*}[h]
	\centering
	\vspace{0mm}
	\hspace{0mm}
	\setlength{\abovecaptionskip}{1mm}
	\setlength{\belowcaptionskip}{1mm}
	\includegraphics[width=0.99\linewidth]{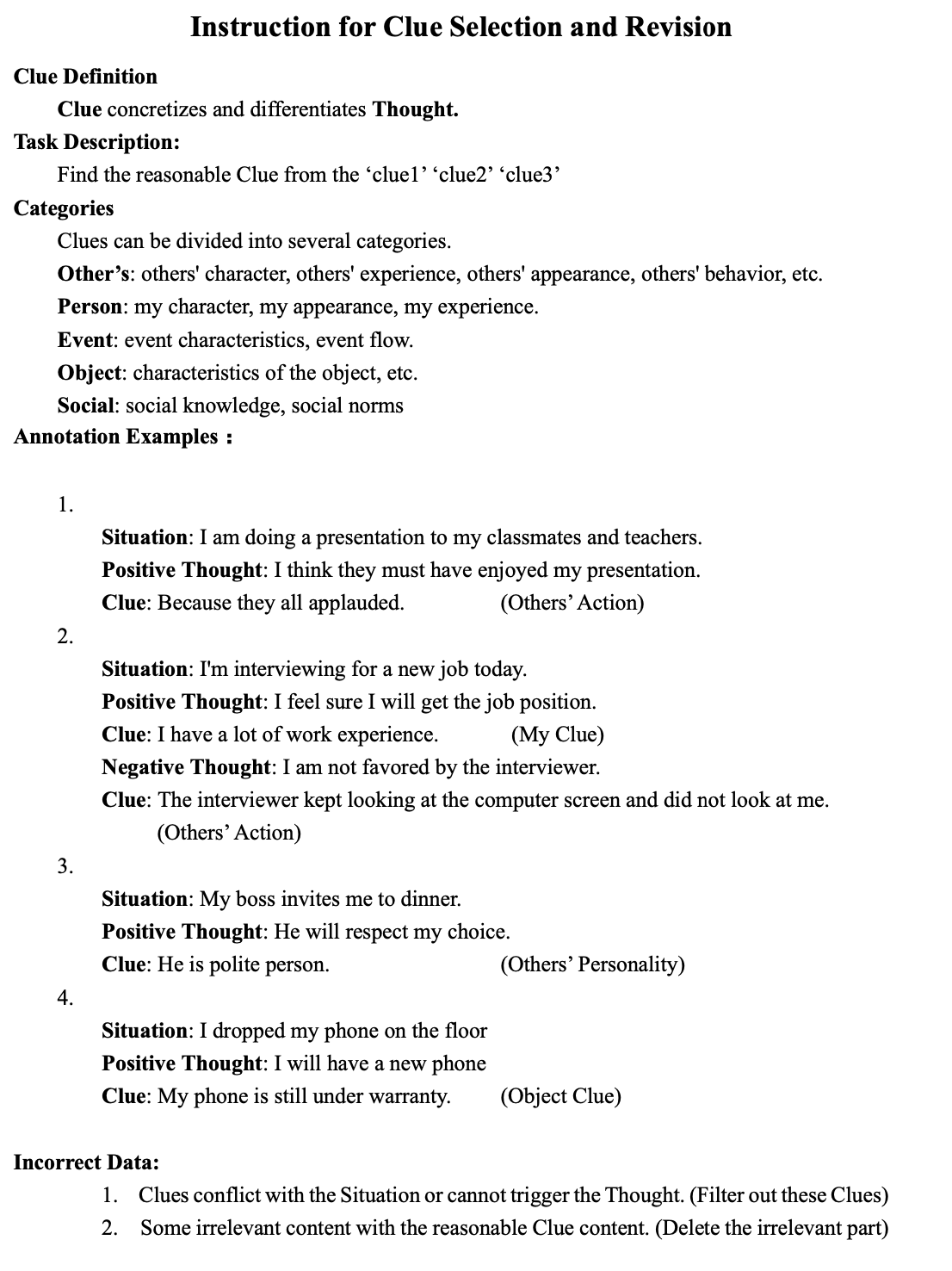}
	\centering
	\caption{Instruction for Clue selection and revision. }
	\label{fig-clue_instruction}
	\vspace{0mm}
\end{figure*}

\begin{figure*}[h]
	\centering
	\vspace{0mm}
	\hspace{0mm}
	\setlength{\abovecaptionskip}{1mm}
	\setlength{\belowcaptionskip}{1mm}
	\includegraphics[width=0.99\linewidth]{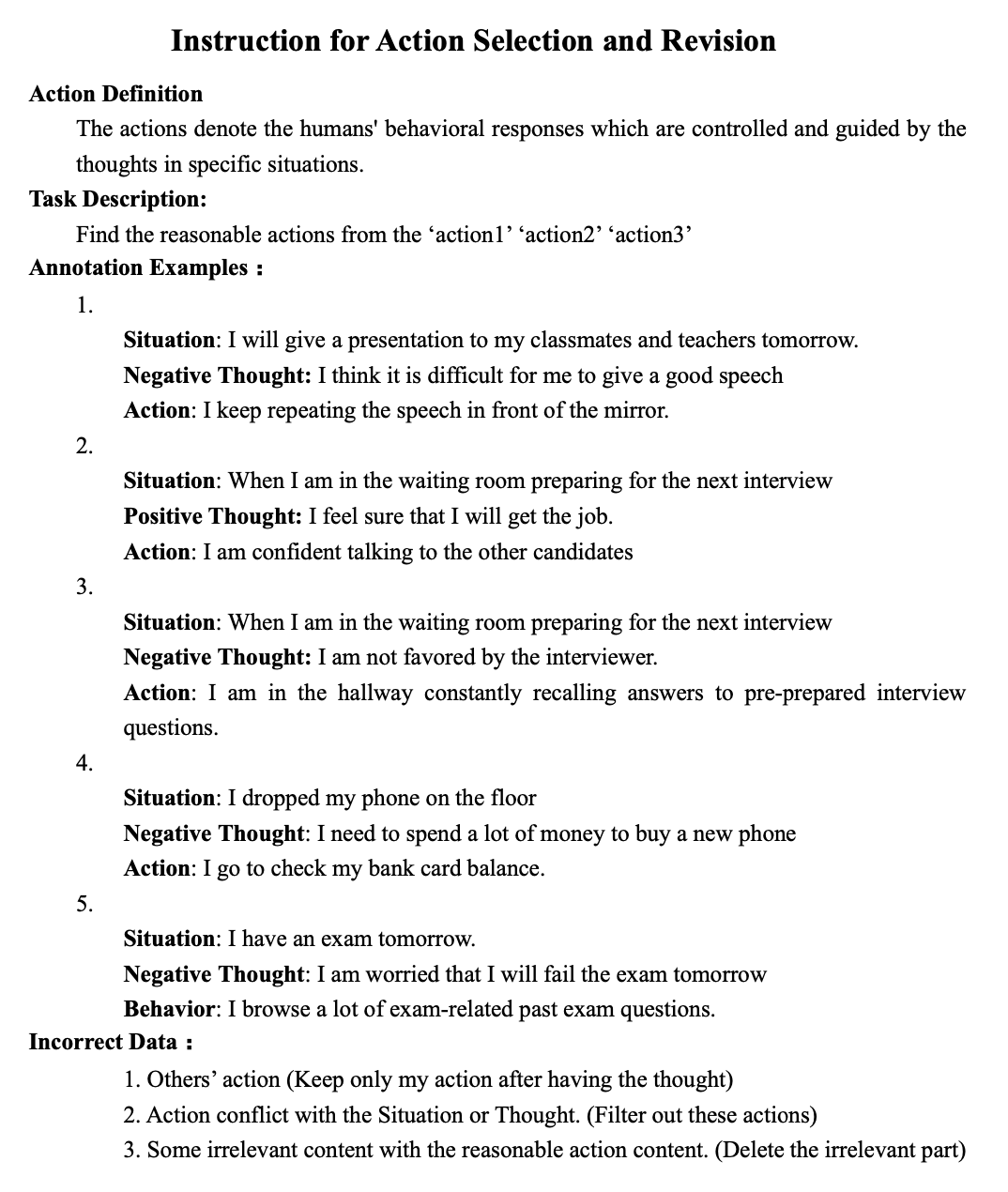}
	\centering
	\caption{Instruction for Action selection and revision. }
	\label{fig-action_instruction}
	\vspace{0mm}
\end{figure*}

\begin{figure*}[h]
	\centering
	\vspace{0mm}
	\hspace{0mm}
	\setlength{\abovecaptionskip}{1mm}
	\setlength{\belowcaptionskip}{1mm}
	\includegraphics[width=0.99\linewidth]{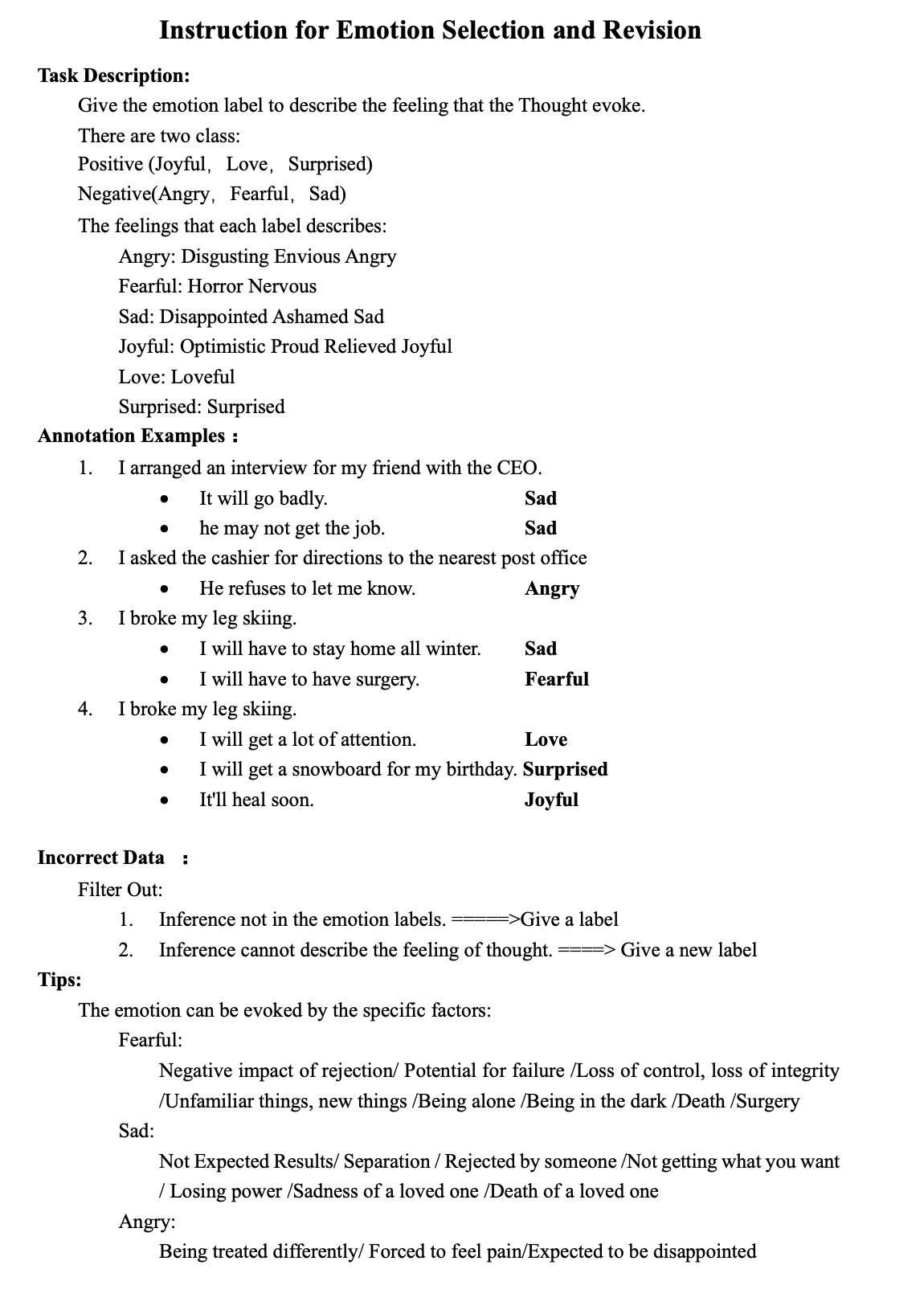}
	\centering
	\caption{Instruction for Emotion selection and revision. }
	\label{fig-emotion_instruction}
	\vspace{0mm}
\end{figure*}

In order to effectively train the annotators, we first have a testing annotation on 160 situations for each task ourselves. Afterwards, we determine what types of incorrect data annotators may encounter and utilize these annotation examples to provide specific guidance. Several common types of bad data in all tasks are 1) repetitive context, 2) unsafe words, and 3) offensive content. Therefore, we automatically filter out the repetitive context and let the annotator manually remove the unsafe terms and offensive content. Besides, the specific types of incorrect data for each task are depicted in the details instructions in Figure \ref{fig-thought_instruction}, Figure \ref{fig-clue_instruction}, Figure \ref{fig-action_instruction}, and Figure \ref{fig-emotion_instruction}. In order to ensure data diversity, we ask workers to modify repetitive data into different and reasonable data.

During training, the annotators work on ten testing situations and receive feedback after following the instructions. Then, based on their annotations of the second 10 situation examples, we evaluate if students have a solid grasp of the task. With specific guidance, students all do well in the subsequent annotation. In addition, to maintain the high quality of the data, we allow them to highlight data with which they are confused and work as the experts to make the ultimate decision. For example, in emotion data revision, the annotator assigns additional emotion labels to the illogical inference, and the experts make the final decision.

\section{Example Cognitive Chains in \coke}
\label{appendix-dataexample}
\begin{figure*}[h]
	\centering
	\vspace{0mm}
	\hspace{0mm}
	\setlength{\abovecaptionskip}{1mm}
	\setlength{\belowcaptionskip}{1mm}
	\includegraphics[width=0.99\linewidth]{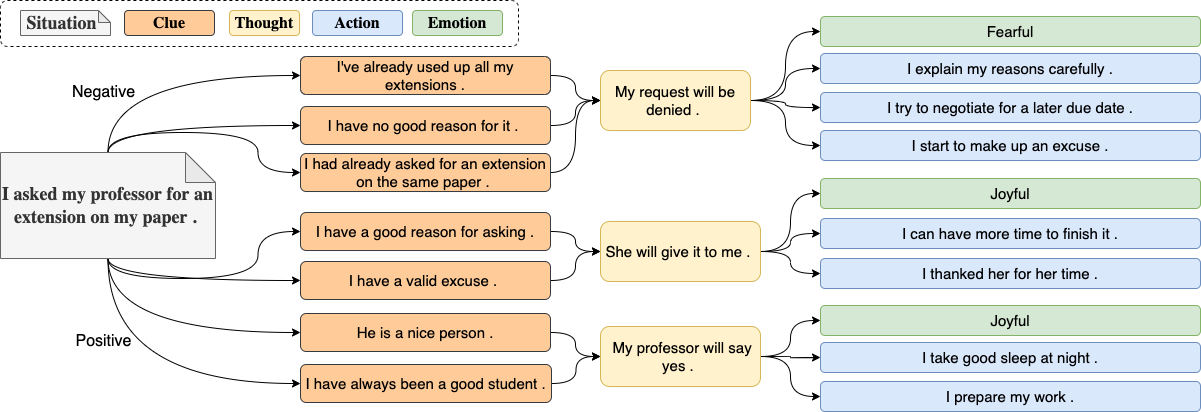}
	\centering
	\caption{Examples of cognitive chains in the situation from the \textbf{School} topic in \coke. }
	\label{fig-CokeExample_School}
	\vspace{20mm}
\end{figure*}

\begin{figure*}[h]
	\centering
	\vspace{0mm}
	\hspace{0mm}
	\setlength{\abovecaptionskip}{1mm}
	\setlength{\belowcaptionskip}{1mm}
	\includegraphics[width=0.99\linewidth]{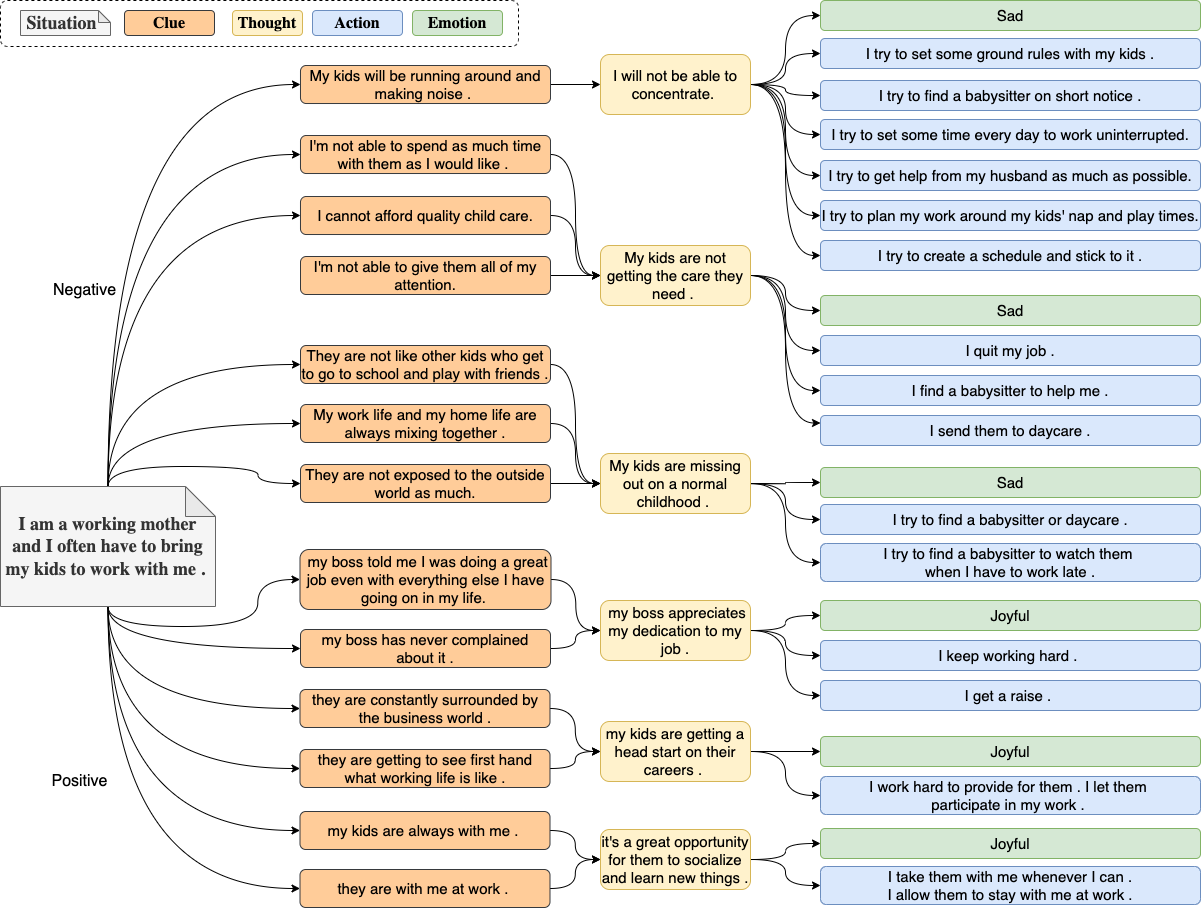}
	\centering
	\caption{Examples of cognitive chains in the situation from the \textbf{Work} topic in \coke. }
	\label{fig-CokeExample_Work}
	\vspace{0mm}
\end{figure*}

\begin{figure*}[h]
	\centering
	\vspace{0mm}
	\hspace{0mm}
	\setlength{\abovecaptionskip}{1mm}
	\setlength{\belowcaptionskip}{1mm}
	\includegraphics[width=0.99\linewidth]{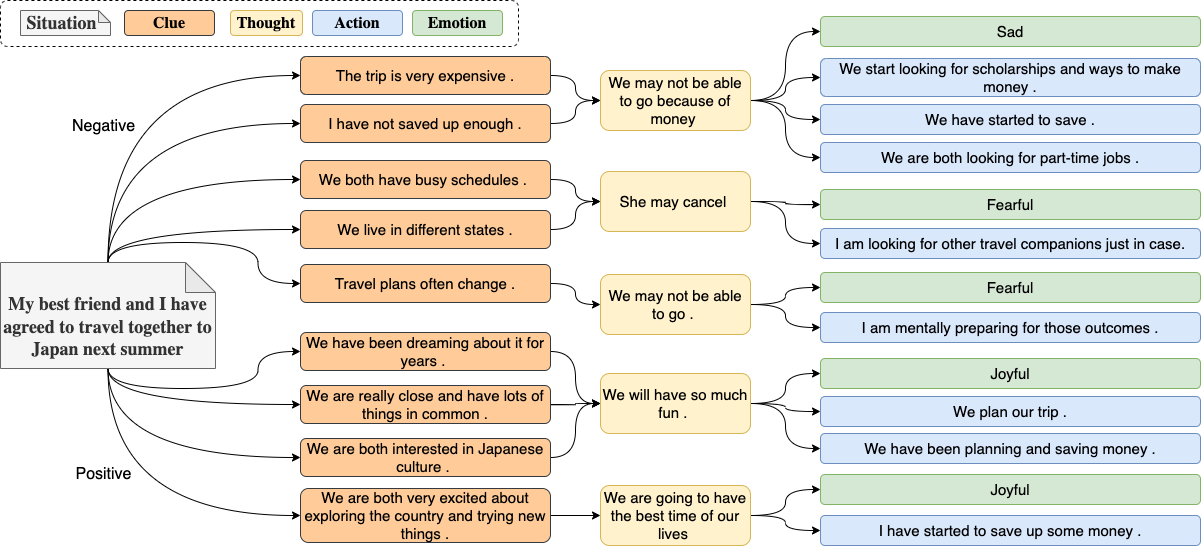}
	\centering
	\caption{Examples of cognitive chains in the situation from the \textbf{Tourism} topic in \coke. }
	\label{fig-CokeExample_Tourism}
	\vspace{20mm}
\end{figure*}

\begin{figure*}[h]
	\centering
	\vspace{0mm}
	\hspace{0mm}
	\setlength{\abovecaptionskip}{1mm}
	\setlength{\belowcaptionskip}{1mm}
	\includegraphics[width=0.99\linewidth]{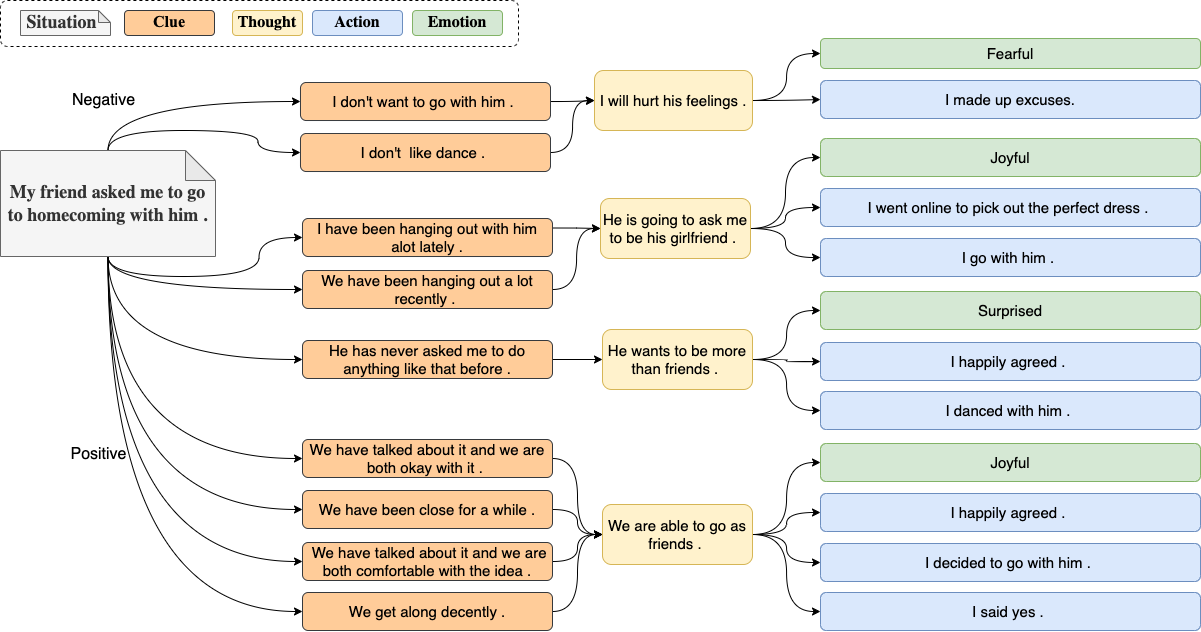}
	\centering
	\caption{Examples of cognitive chains in the situation from the \textbf{Relationship} topic in \coke. }
	\label{fig-CokeExample_Relationship}
	\vspace{0mm}
\end{figure*}

\begin{figure*}[h]
	\centering
	\vspace{0mm}
	\hspace{0mm}
	\setlength{\abovecaptionskip}{1mm}
	\setlength{\belowcaptionskip}{1mm}
	\includegraphics[width=0.99\linewidth]{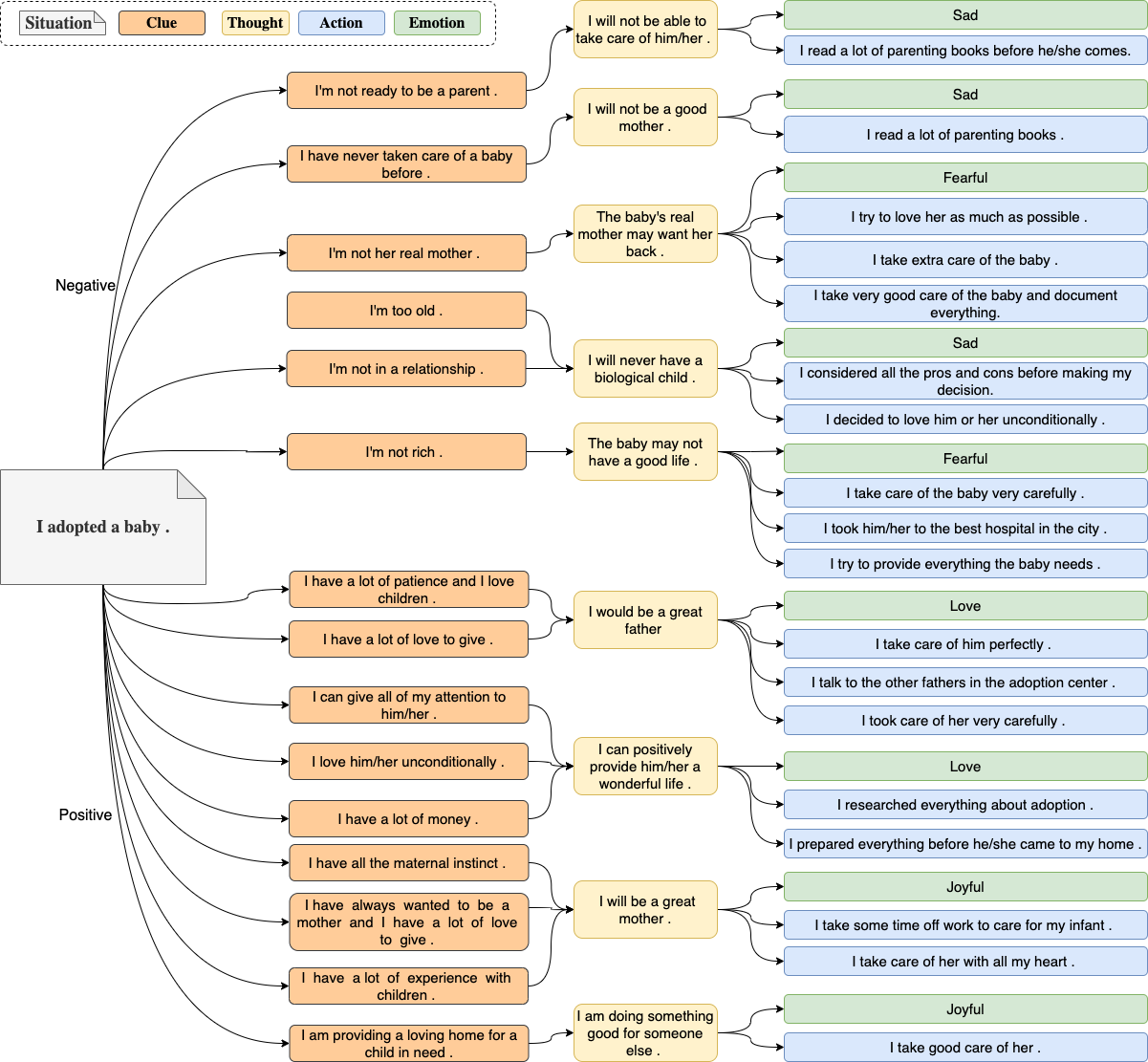}
	\centering
	\caption{Examples of cognitive chains in the situation from the \textbf{Ordinary Life} topic in \coke. }
	\label{fig-CokeExample_Life}
	\vspace{0mm}
\end{figure*}

We present the cognitive chains in \coke from five topics in Figure \ref{fig-CokeExample_School} (School), Figure \ref{fig-CokeExample_Work} (Work), Figure \ref{fig-CokeExample_Tourism} (Tourism), Figure \ref{fig-CokeExample_Relationship} (Relationship), and Figure \ref{fig-CokeExample_Life} (Ordinary Life), respectively. The data in \coke is in English.

\section{Prompts for PLMs Generation in Evaluation }
\label{experiment-prompt}
As shown in Table \ref{tab-T5_test }, we present the prompts that lead the LLaMA-2, Mistral, GPT-3.5 Turbo and GPT-4 to make inferences on the validation dataset.

\input{tables/AppendixE_test_prompt.tex}

\section{Case Study of Generated Cognitive Chains}
\label{appendix-casestudy}
To have a close look, in Table \ref{tab-casestudy}, we present ten cognitive chains generated by \cokelm in five situations covering different topics. It can be observed that the proposed cognitive generation model \cokelm can generate smooth and effective cognitive chains in unseen situations.

\input{tables/case_study.tex}

\section{Empowering Emotional Support Conversation with \coke }
\label{appendix-esc}

In this section, we discuss how \coke can be used to enhance social applications, specifically focusing on Emotional Support Conversation (ESC) \cite{DBLP:conf/acl/LiuZDSLYJH20}. This task represents a key arena in which Theory of Mind (ToM) can play a pivotal role, given the need to understand and respond to a user's emotional state and underlying challenges.

\paragraph{Emotional Support Conversation}

The ESC task is structured around a user in a negative emotional state, potentially due to a specific problem or challenge they are confronting. The user's emotional state is characterized by a negative emotion label (e), an intensity level of the emotion (l, on a scale of 1 to 5), and the underlying challenge causing their distress. The goal of the supporter (or AI system) is to comfort the user in a conversation, deploying support skills to reduce the intensity of the user's negative emotions. The conversation's effectiveness is gauged by the extent to which the user's emotional intensity is reduced, and the ability of the supporter to accurately identify the problem, comfort the user, and offer constructive solutions or suggestions.

\paragraph{Experimental Results}
In experiments, we compare three models: Vanilla ESC,  ESC w/ \cokelm Actions, and ESC w/ \cokelm Thoughts. We conduct both automatic evaluation (PPL, BLEU-2, ROUGE-L, Extrema) and human evaluation (Fluency, Indentification, Comforting, Suggestion), same as the original ESC paper. The experimental results are presented in Table \ref{tab-esc-autoevaluation}. To have a close look, we also present a case study in Table \ref{tab-esc-case}. The results demonstrate the potential of \coke to empower social applications, offering a more nuanced and empathetic approach to AI-driven emotional support. By integrating \coke's cognitive chains into dialogue history, AI systems can achieve a more sophisticated understanding of human emotions and behavior, leading to more effective interactions in emotionally charged contexts such as the ESC task. Interestingly, the ESC with \cokelm Thoughts model shows a slight advantage over the ESC with \cokelm Actions model. This could be attributed to the fact that the thoughts generated by \cokelm offer a more direct reflection of the user's mental state.

\input{tables/AppendixH_Example}

\begin{figure*}[h]
	\centering
	\vspace{1mm}
	\hspace{-2mm}
	\setlength{\abovecaptionskip}{0mm}
	\setlength{\belowcaptionskip}{0mm}
        \includegraphics[width=0.98\linewidth]{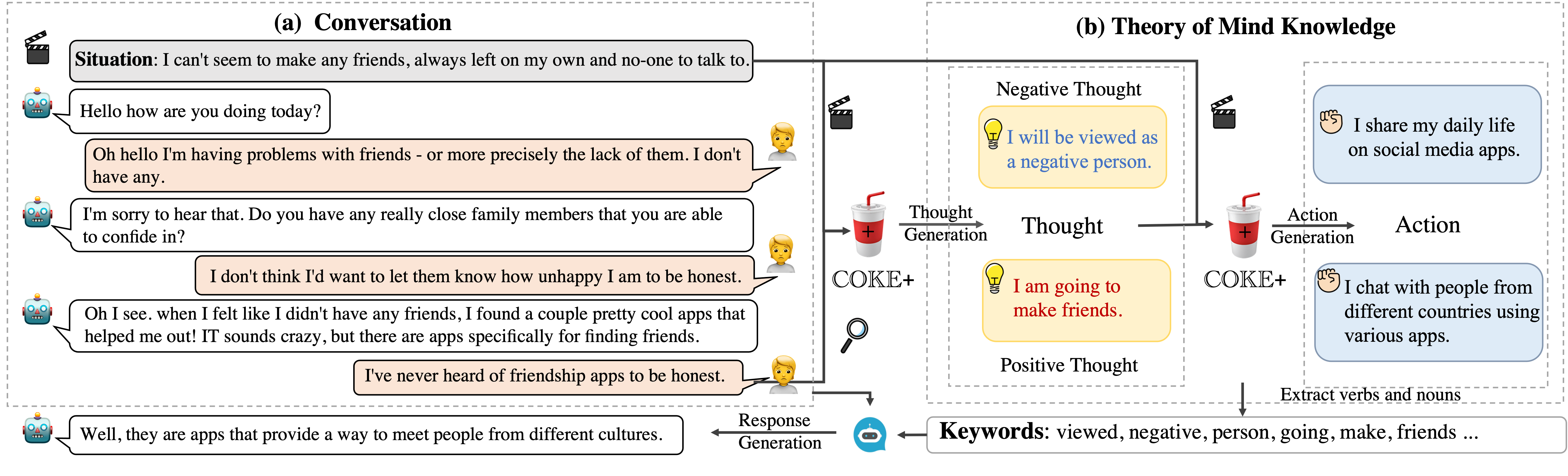}
	\centering
        \caption{Empowering ESC with \cokelm. A detailed case study is illustrated in Table \ref{tab-esc-case}. }
	\label{fig-empowering}
	\vspace{-2mm}
\end{figure*}





\input{tables/AppendixB_prompt_topic.tex}

\clearpage

\clearpage

\clearpage

\clearpage

\clearpage

\clearpage

%% file: tables/AppendixA_keywords.tex
\begin{table}[h]
\centering
\scalebox{0.82}{
\begin{tabularx}{.56\textwidth}{
			p{.11\textwidth} <{\raggedright}
               p{.4\textwidth} <{\raggedright}}
               
\toprule
\textbf{Topic} & \multicolumn{1}{l}{\textbf{Keywords extracted}}                               \\ \midrule
School         & asked, friend, friends, dad, project, mom, help, go  \\ \midrule
Work           & boss, ask, work, asked,  new, friend, job, meeting         \\ \midrule
Tourism         & asked, trip, travel, go, friend, friends, sister,   brother \\ \midrule
Relationship   & asked, new, date, friends, best, girl,guy, girlfriend       \\ \midrule
Ordinary Life  & asked,day, friend, new, friends, go, wanted, going  \\ \bottomrule
\end{tabularx}}

\caption{Keywords in five topics. }
\label{tab-keywords}
\vspace{0mm}
\end{table}

%% file: tables/AppendixD_GPT3.tex
\begin{table}[h]
\vspace{2mm}
\centering
\scalebox{0.68}{
\begin{tabularx}{.68\textwidth}{
			p{.16\textwidth} <{\raggedright}
               p{.08\textwidth} <{\centering}
               p{.06\textwidth} <{\centering}
               p{.07\textwidth} <{\centering}
                p{.06\textwidth} <{\centering}
			p{.07\textwidth} <{\centering}} 
\toprule
\textbf{Parameter} & \textbf{Situation} & \textbf{Clue } & \textbf{Thought} & \textbf{Action} & \textbf{Emotion} \\ \midrule
\textbf{n}                  & 1   & 3   & 3   & 3   & 1   \\ \midrule
\textbf{best\_of}           & 1   & 3   & 3   & 3   & 1   \\ \midrule
\textbf{model}     & text-davinci-002              & text-davinci-002         & text-davinci-002            & text-davinci-002           & text-davinci-002            \\ \midrule
\textbf{temperature}        & 1   & 1   & 1   & 1   & 1   \\ \midrule
\textbf{max\_tokens}        & 256 & 256 & 256 & 256 & 256 \\ \midrule
\textbf{top\_p}             & 1   & 1   & 1   & 1   & 1   \\ \midrule
\textbf{frequency\_penalty} & 0   & 0   & 0   & 0   & 0   \\ \midrule
\textbf{presence\_penalty}  & 0   & 0   & 0   & 0   & 0   \\ \bottomrule
\end{tabularx}%

}
\caption{Parameters for GPT-3.5 generation.}
\label{tab-IntructGPT_parameter}
\vspace{2mm}
\end{table}

%% file: tables/AppendixE_test_prompt.tex
\begin{table*}[h]
\caption{ Prompts for LLaMa/Mistral/GPT-3.5-Turbo/GPT-4 generation on the validation data in the negative and positive cognitive chains.}
\vspace{0mm}
\small
\renewcommand\arraystretch{1.2}
\setlength{\abovecaptionskip}{1mm}
\setlength{\belowcaptionskip}{1mm}
\centering
\setlength{\tabcolsep}{0.7mm}
\resizebox{\textwidth}{!}{
\begin{tabular}{@{}ccl@{}}
\toprule
\toprule
\textbf{Type}                                                         & \textbf{} & \multicolumn{1}{c}{\textbf{Prompt for testing clue   collection}}                   \\ \midrule
\multicolumn{1}{l}{Clue generation for the negative cognitive chain} & L1        & Complete the sentence with the negative clue:  \\
                                                                      & L2       & When Situation, I think negatively since   \\
\multicolumn{1}{l}{Clue generation for the positive cognitive chain} & L1        & Complete the sentence with the positive clue:  \\
                                                                      & L2       & When Situation, I think positively since   \\
\toprule
\toprule
\textbf{Type}                                                         & \textbf{} & \multicolumn{1}{c}{\textbf{Prompt for testing though   collection}}                 \\ \midrule
\multicolumn{1}{l}{Thought  generation for the negative cognitive chain} & L1 & Complete the sentence with the negative thought: \\
                                                                         & L2 & When Situation and Clue, I feel terrible since I think \\
\multicolumn{1}{l}{Thought  generation for the positive cognitive chain} & L1 & Complete the sentence with the positive thought: \\
                                                                         & L2 & When Situation and Clue, I feel great since I think   \\
\toprule
\toprule
\textbf{Type}                                                         & \textbf{} & \multicolumn{1}{c}{\textbf{Prompt for testing action   collection}}                 \\ \midrule
\multicolumn{1}{l}{Action generation for the negative cognitive chain} & L1 & Complete the sentence with the negative action: \\
                                                                       & L2 & When Situation, I think Thought, so  \\
\multicolumn{1}{l}{Action generation for the positive cognitive chain} & L1 & Complete the sentence with the positive action: \\
                                                                       & L2 & When Situation, I think Thought, so \\ 
\toprule
\toprule
\textbf{Type}                                                         & \textbf{} & \multicolumn{1}{c}{\textbf{Prompt for testing emotion   collection}}                \\ \midrule
\multirow{2}{*}{Emotion generation for the negative cognitive chain} & L1 & Choose one word from Sad, Angry, Fearful to describe the given situation:          \\
                                                                     & L2 & When Situation, I think Thought. \\
\multirow{2}{*}{Emotion generation for the positive cognitive chain} & L1 & Choose one word from Love, Surprise, Joyful to describe the given situation:       \\
                                                                     & L2 & When Situation, I think Thought.  \\                   \bottomrule
\end{tabular}%
}
\label{tab-T5_test }
\vspace{0mm}
\end{table*}

%% file: tables/case_study.tex
\begin{table*}[h]
\caption{Case study. Given situations from different topics, the proposed cognitive generation model \cokelm can predict corresponding clues, thoughts, actions, and emotions in a pipeline manner.}
\centering
\scalebox{0.6}{
\begin{tabularx}{1.65\textwidth}{
			      p{.14\textwidth} <{\raggedright}
                    p{.25\textwidth} <{\raggedright}
                    p{.03\textwidth} <{\raggedright}
                    p{.32\textwidth} <{\raggedright}
                    p{.32\textwidth} <{\raggedright}
                    p{.32\textwidth} <{\raggedright}
		        p{.08\textwidth} <{\centering}}

\toprule
\textbf{Topic} & \textbf{Situation} & & \textbf{Clue} & \textbf{Thought} & \textbf{Action} & \textbf{Emotion}
\\
\midrule
 &
   &
    \texttt{Neg.}&
  It was worth more than my car. &
  The phone is going to be ruined. &
  I quickly sent an email to my boss to own up to the mistake. &
  Fearful\\ \cmidrule(l){3-7} 
\multirow{-3}{*}{Work} &
  \multirow{-3}{*}{\begin{tabular}[c]{@{}l@{}}I accidentally dropped my\\ companies latest phone \\prototype.\end{tabular}} &
   \texttt{Pos.}&
  I have been working so hard on this project. &
  My boss will forgive me. &
  I quickly picked it up and acted like it was no big deal. &
  Joyful \\ \midrule
 &
   &
   \texttt{Neg.}&
  He has been acting distant recently. &
  He may not show up.  &
  I asked for some friends as backup. &
  Sad \\ \cmidrule(l){3-7} 
\multirow{-2}{*}{Relationship} &
  \multirow{-2}{*}{\begin{tabular}[c]{@{}l@{}}I also set up a romantic \\ evening for the two of us.\end{tabular}} &
   \texttt{Pos.}&
  We have been getting along very well lately. &
  He is going to propose to me. &
  I am going to take him to his favorite restaurant. &
  Love \\ \midrule
 &
   &
  \texttt{Neg.}&
  I have never done things differently before. &
  I will not be able to understand the material.  &
  I read through the material again. &
  Fearful \\ \cmidrule(l){3-7} 
\multirow{-2}{*}{School} &
  \multirow{-2}{*}{\begin{tabular}[c]{@{}l@{}}I am going to  adopt a new\\ approach to my studies.\end{tabular}} &
   \texttt{Pos.}&
  I have put in a lot of effort. &
  It will improve my grades.  &
  I take my study time more seriously.  &
  Joyful \\ \midrule
 &
   &
  \texttt{Neg.}&
  I do not pack enough money. &
  I will regret not spending my time more wisely.&
  I asked him to take me to as many places as he could. &
  Sad \\ \cmidrule(l){3-7} 
\multirow{-2}{*}{Tourism} &
  \multirow{-2}{*}{\begin{tabular}[c]{@{}l@{}}I will accompany my \\ brother to Greece for \\the summer.\end{tabular}} &
   \texttt{Pos.}&
  We explored different cities and towns. &
  I will have a great time there. &
  I pack my bags with the best clothes I have.  &
  Joyful \\ \midrule
 &
   &
   \texttt{Neg.}&
  He is not my real father. &
  He will spend all my money.  &
  I find a more reliable person. &
  Fearful \\ \cmidrule(l){3-7} 
\multirow{-2}{*}{Ordinary Life} &
  \multirow{-2}{*}{\begin{tabular}[c]{@{}l@{}}My dad acts as my power \\of attorney .\end{tabular}} &
   \texttt{Pos.}&
  My dad is really responsible and always helps me out. &
  He will make good decisions on my behalf.  &
  I follow his instructions.  &
  Joyful \\ \bottomrule
\end{tabularx}%

}
\label{tab-casestudy}
\vspace{0mm}
\end{table*}

%% file: tables/AppendixH_Example.tex
\begin{table*}[h]
\vspace{-6mm}
\caption{Case study of emotional support conversation.}
\footnotesize
\centering
\renewcommand\arraystretch{1.20}
\resizebox{0.95\textwidth}{!}{%
\begin{tabular}{m{4cm} m{14cm}}
\\\hline
\multirow{15}{*}{Dialogue History} &\colorbox{situationcolor}{SITUATION:  I can't seem to make any friends - always left on my own and no-one to talk to.}  \\
& {SYSTEM:  Hello how are you doing today? I'd love to chat with you! What's been on your mind?}  \\
 & \colorbox{cluecolor}{USER:  oh hello I'm having problems with friends - or more precisely the lack of them. I don't have any.}  \\
& {SYSTEM:  Oh I see. You feel as though you don't have any friend's currently? I totally get that. I feel like that a lot too, especially in these weird times we are going through. It's so hard to see people and keep relationships right now}  \\
& \colorbox{cluecolor}{USER:  Well it's more that I think I've never had friends. Not anyone I can confide in or laugh with} \\
& {SYSTEM:  That's a really hard thing to be going through. I'm sorry to hear that. Do you have any really close family members that you are able to confide in?} \\
& \colorbox{cluecolor}{USER:  I don't think I'd want to let them know how unhappy I am to be honest} \\
& {SYSTEM:  Oh I see. Well personally, when I was going through a hard time and I felt like I didn't have any friends, I found a couple pretty cool apps that helped me out! IT sounds crazy, but there are apps specifically for finding friends. They are kind of like dating apps but it's 100\% meant for friends so everyone on them has the same intentions you do. I actually found a couple close friends through apps like that. Have you heard about those or looked into them? }\\
& \colorbox{cluecolor}{USER:  I've never heard of friendship apps to be honest} \\ \hline
Keywords   & \colorbox{thoughtcolor}{viewed,negative,person,going,confront,something, alone,need,make,friends} \\ \hline
Golden Response  & SYSTEM:  Yeah! They are a really good way to connect with people that you share a lot of similar interests with. \\ \hline
Vanilla ESC   & SYSTEM:  it's a very popular app and there are tons of sites like facetime and facetime.  \\ \hline
ESC w/ \cokelm Thoughts & SYSTEM:  well, they are apps that provide a way to meet people from different cultures.  \\ \hline                                                                                                                                                                                                                                                                                                                                                                                             
\end{tabular}
}
\label{tab-esc-case}
\end{table*}

%% file: tables/AppendixB_prompt_topic.tex
\begin{table*}[htbp]
\caption{Prompts for collecting \textbf{Situations} under five common daily topics. Each prompt input consists of 6 lines of content. The first line \textcolor[rgb]{0.502,0.502,0.502}{L1} is an \textbf{Introduction}, lines \textcolor[rgb]{0.502,0.502,0.502}{L2} - \textcolor[rgb]{0.502,0.502,0.502}{L5} are \textbf{Demonstration}, and the last line \textcolor[rgb]{0.502,0.502,0.502}{L6} is \textbf{Query}.}
\centering
\renewcommand\arraystretch{1.5}
\scalebox{0.68}{
\begin{tabularx}{1.45\textwidth}{
p{.15\textwidth} <{\centering}
p{.02\textwidth} <{\raggedleft}
p{1.2\textwidth} <{\raggedright}}

\toprule
\textbf{Topic} &   & \textbf{Prompt inputs for collecting \emph{Situations}}
\\\midrule

\multirow{6}{*}{\begin{tabular}[c]{@{}c@{}}(a) \\School\end{tabular}}        & \textcolor[rgb]{0.502,0.502,0.502}{L1} & Translate [Sentence] to [Situation]:                                                                                                                                                                                                   \\
& \textcolor[rgb]{0.502,0.502,0.502}{L2} & {[}Sentence] \textcolor{myblue}{PersonX gives a presentation.} =\textgreater{} {[}Situation] \textcolor{myorange}{I will give a presentation at college tomorrow .}                                                                \\
& \textcolor[rgb]{0.502,0.502,0.502}{L3} & {[}Sentence] \textcolor{myblue}{PersonX has PersonX's driving test.} =\textgreater{} {[}Situation] \textcolor{myorange}{My mom has her driving test tomorrow.~}                                                                    \\
& \textcolor[rgb]{0.502,0.502,0.502}{L4} & {[}Sentence] \textcolor{myblue}{PersonX gives PersonY a book. }=\textgreater{} {[}Situation] \textcolor{myorange}{My professor give me a book for research.~}                                                                      \\
& \textcolor[rgb]{0.502,0.502,0.502}{L5} & {[}Sentence] \textcolor{myblue}{PersonX pays PersonY's fees.} =\textgreater{} {[}Situation] \textcolor{myorange}{My aunt paid my test fees.~}                                                                                      \\
& \textcolor[rgb]{0.502,0.502,0.502}{L6} & {[}Sentence] \textcolor{myblue}{\uline{Selected Atomic Events}} =\textgreater{} {[}Situation]                                                                                                                                          \\ 
\midrule
\multirow{6}{*}{\begin{tabular}[c]{@{}c@{}}(b) \\Work\end{tabular}}          & \textcolor[rgb]{0.502,0.502,0.502}{L1} & Translate [Sentence] to [Situation]:~                                                                                                                                                                                                  \\
& \textcolor[rgb]{0.502,0.502,0.502}{L2} & {[}Sentence] \textcolor{myblue}{PersonX has a meeting .} =\textgreater{} {[}Situation]\textcolor{myorange}{ I will have a meeting with my boss .}                                                                                  \\
& \textcolor[rgb]{0.502,0.502,0.502}{L3} & {[}Sentence] \textcolor{myblue}{PersonX took PersonY .} =\textgreater{} {[}Situation] \textcolor{myorange}{I went out to eat with my boss, and he took me to a really nice restaurant.~}                                           \\
& \textcolor[rgb]{0.502,0.502,0.502}{L4} & {[}Sentence] \textcolor{myblue}{PersonX announces PersonX's decision .} =\textgreater{} {[}Situation] \textcolor{myorange}{I announce my decision to accept a~ job position that I did not want in order to please my father .}    \\
& \textcolor[rgb]{0.502,0.502,0.502}{L5} & {[}Sentence] \textcolor{myblue}{PersonX becomes PersonY's~ .} =\textgreater{} {[}Situation] \textcolor{myorange}{My friend becomes my boss.~}                                                                                      \\
& \textcolor[rgb]{0.502,0.502,0.502}{L6} & {[}Sentence] \textcolor{myblue}{\uline{Selected Atomic Events}} =\textgreater{} {[}Sentence]~                                                                                                                                            \\ 
\midrule
\multirow{6}{*}{\begin{tabular}[c]{@{}c@{}}(c) \\Ordinary Life\end{tabular}} & \textcolor[rgb]{0.502,0.502,0.502}{L1} & Translate [Sentence] to [Situation]:~                                                                                                                                                                                                  \\
& \textcolor[rgb]{0.502,0.502,0.502}{L2} & {[}Sentence] \textcolor{myblue}{PersonX watches the game .} =\textgreater{} {[}Situation] \textcolor{myorange}{I watched Wimbeldon for the first time .}                                                                           \\
& \textcolor[rgb]{0.502,0.502,0.502}{L3} & {[}Sentence] \textcolor{myblue}{PersonX gives PersonY advice .} =\textgreater{} {[}Situation] \textcolor{myorange}{My friend gives me financial advice on stocks .~}                                                               \\
& \textcolor[rgb]{0.502,0.502,0.502}{L4} & {[}Sentence] \textcolor{myblue}{PersonX visits PersonX's friends .} =\textgreater{} {[}Situation] \textcolor{myorange}{I went to Cuba to visit some family members and met up with some friends to go to a bar .~}                 \\
& \textcolor[rgb]{0.502,0.502,0.502}{L5} & {[}Sentence] \textcolor{myblue}{PersonX is cleaning PersonY's house . }=\textgreater{} {[}Situation] \textcolor{myorange}{I was helping clean out my parents' house the other day and found all my old high school year books .~}  \\
& \textcolor[rgb]{0.502,0.502,0.502}{L6} & {[}Sentence] \textcolor{myblue}{\uline{Selected Atomic Events }}=\textgreater{} {[}Situation]~                                                                                                                                           \\ 
\midrule
\multirow{6}{*}{\begin{tabular}[c]{@{}c@{}}(d) \\Tourism\end{tabular}}       & \textcolor[rgb]{0.502,0.502,0.502}{L1} & Translate [Sentence] to [Situation]:~                                                                                                                                                                                                  \\
& \textcolor[rgb]{0.502,0.502,0.502}{L2} & {[}Sentence] \textcolor{myblue}{PersonX travels .} =\textgreater{} {[}Situation]\textcolor{myorange}{ My aunt traveled all by herself .~}                                                                                          \\
& \textcolor[rgb]{0.502,0.502,0.502}{L3} & {[}Sentence] \textcolor{myblue}{PersonX takes PersonY to work .} =\textgreater{} {[}Situation] \textcolor{myorange}{My manager has taken my coworker for a work trip to New York.~}                                                \\
& \textcolor[rgb]{0.502,0.502,0.502}{L4} & {[}Sentence] \textcolor{myblue}{PersonX plans PersonX's next trip . }=\textgreater{} {[}Situation] \textcolor{myorange}{We are planning our first cruise .~}                                                                       \\
& \textcolor[rgb]{0.502,0.502,0.502}{L5} & {[}Sentence] \textcolor{myblue}{PersonX sees PersonY's .} =\textgreater{} {[}Situation]\textcolor{myorange}{ I have seen some pictures of friends of mine traveling around Italy.}                                                 \\
& \textcolor[rgb]{0.502,0.502,0.502}{L6} & {[}Sentence] \textcolor{myblue}{\uline{Selected Atomic Events}} =\textgreater{} {[}Situation]~                                                                                                                                           \\ 
\midrule
\multirow{6}{*}{(e) Relationship}                                            & \textcolor[rgb]{0.502,0.502,0.502}{L1} & ~Translate [Sentence] to [Situation]:~                                                                                                                                                                                                 \\
& \textcolor[rgb]{0.502,0.502,0.502}{L2} & {[}Sentence] \textcolor{myblue}{PersonX gets married}~ =\textgreater{} {[}Situation] \textcolor{myorange}{One of old coworkers is getting married .~}                                                                              \\
& \textcolor[rgb]{0.502,0.502,0.502}{L3} & {[}Sentence] \textcolor{myblue}{PersonX gives~ PersonY .} =\textgreater{} {[}Situation] \textcolor{myorange}{I gave a house key to a girl i had went on 2 dates with .~}                                                           \\
& \textcolor[rgb]{0.502,0.502,0.502}{L4} & {[}Sentence] \textcolor{myblue}{PersonX is on PersonX’s .} =\textgreater{} {[}Situation] \textcolor{myorange}{I am on my first blind date .~}                                                                                      \\
& \textcolor[rgb]{0.502,0.502,0.502}{L5} & {[}Sentence] \textcolor{myblue}{PersonX gets PersonY's number . }=\textgreater{} {[}Situation] \textcolor{myorange}{I got the new girl's number at school .~}                                                                      \\
& \textcolor[rgb]{0.502,0.502,0.502}{L6} & {[}Sentence] \textcolor{myblue}{\uline{Selected Atomic Events}} =\textgreater{} {[}Situation]                                                                                                                                           \\
\bottomrule

\end{tabularx}}

\label{tab-promptforsituation}
\end{table*}

\clearpage

\begin{table*}[htbp]
\caption{Prompts for collecting \textbf{Thoughts}, \textbf{Clues} and \textbf{Actions} in negative and positive cognitive chains. Each prompt input consists of 5 lines of content. The first line \textcolor[rgb]{0.502,0.502,0.502}{L1} is an \textbf{Introduction}, lines \textcolor[rgb]{0.502,0.502,0.502}{L2} - \textcolor[rgb]{0.502,0.502,0.502}{L4} are \textbf{Demonstration}, and the last line \textcolor[rgb]{0.502,0.502,0.502}{L5} is \textbf{Query}.}
\centering
\renewcommand\arraystretch{1.5}
\scalebox{0.68}{
\begin{tabularx}{1.45\textwidth}{
p{.15\textwidth} <{\centering}
p{.02\textwidth} <{\raggedleft}
p{1.2\textwidth} <{\raggedright}}
\toprule
\toprule
\textbf{Type} &   & \textbf{Prompt inputs for collecting \emph{Thoughts}}
\\\midrule

\multirow{5}{*}{\begin{tabular}[c]{@{}c@{}}(a) \\Thought \\in the negative \\cognitive Chain~~\end{tabular}} & \textcolor[rgb]{0.502,0.502,0.502}{L1} & Complete the sentence:                                                                                                                                                         \\
 & \textcolor[rgb]{0.502,0.502,0.502}{L2} & When\textcolor{myblue}{ I will give a presentation at college}, I feel terrible since I think \textcolor{myorange}{I will freeze on stage.~}\textcolor{myblue}{~}            \\
 & \textcolor[rgb]{0.502,0.502,0.502}{L3} & When\textcolor{myblue}{ I was waiting in the waiting room for my last job interview,} I felt terrible since I thought\textcolor{myorange}{ I would not be selected.}~      \\
 & \textcolor[rgb]{0.502,0.502,0.502}{L4} & When \textcolor{myblue}{I lost the wallet}, I felt terrible since I thought\textcolor{myorange}{ I would be blamed}.~                                                      \\
 & \textcolor[rgb]{0.502,0.502,0.502}{L5} & When \textcolor{myblue}{\uline{Situation}}, I feel terrible since I think~                                                                                                       \\ 

\midrule
\multirow{5}{*}{\begin{tabular}[c]{@{}c@{}}(b) \\Thought \\in the positive\\~cognitive chain~~\end{tabular}} & \textcolor[rgb]{0.502,0.502,0.502}{L1} & Complete the sentence:~                                                                                                                                                        \\
 & \textcolor[rgb]{0.502,0.502,0.502}{L2} & When \textcolor{myblue}{I will give a presentation at college}, I feel great since I think \textcolor{myorange}{ I am going to ace this presentation.~}                     \\
 & \textcolor[rgb]{0.502,0.502,0.502}{L3} & When \textcolor{myblue}{I was waiting in the waiting room for my last job interview,} I felt great since I thought \textcolor{myorange}{I would land my dream job.~}       \\
 & \textcolor[rgb]{0.502,0.502,0.502}{L4} & When \textcolor{myblue}{I gave a house key to the girl I had gone on two dates with}, I felt great since I thought \textcolor{myorange}{she would fall in love with me.~}  \\
 & \textcolor[rgb]{0.502,0.502,0.502}{L5} & When \textcolor{myblue}{\uline{Situation}}, I feel great since I think    \\
\bottomrule
\toprule
\textbf{Type}                                                                                                     &                                        & \textbf{Prompt inputs for collecting \textit{Clues}}                                                                                                                                                                                                                      \\ 
\midrule
\multirow{5}{*}{\begin{tabular}[c]{@{}c@{}}(a)\\Clue\\in the negative \\cognitive chain~\end{tabular}}     & \textcolor[rgb]{0.502,0.502,0.502}{L1} & Complete the sentence:                                                                                                                                                                                                                                                    \\
          & \textcolor[rgb]{0.502,0.502,0.502}{L2} & When \textcolor{myblue}{I will give a presentation at college}, I think\textcolor{myblue}{ I will freeze on stage}\textcolor{myorange}{ }since\textcolor{myorange}{ I haven't given a presentation before.~}                                                      \\
          & \textcolor[rgb]{0.502,0.502,0.502}{L3} & When \textcolor{myblue}{I was waiting in the waiting room for my last job interview}, I thought\textcolor{myblue}{ I would not be selected }since\textcolor{myorange}{ too many people were interviewing for the job.~}                                                 \\
          & \textcolor[rgb]{0.502,0.502,0.502}{L4} & When \textcolor{myblue}{my college entrance exam is tomorrow}, I think\textcolor{myblue}{ I will fail the exam}\textcolor{myorange}{ }since\textcolor{myorange}{ I didn't prepare well.~~}                                                                        \\
          & \textcolor[rgb]{0.502,0.502,0.502}{L5} & When \textcolor{myblue}{\uline{Situation}}, I think \textcolor{myblue}{\uline{Thought}} since~                                                                                                                                                                                \\ 
\midrule
\multirow{5}{*}{\begin{tabular}[c]{@{}c@{}}(b)\\ Clue \\in the positive \\cognitive chain~\end{tabular}}     & \textcolor[rgb]{0.502,0.502,0.502}{L1} & Complete the sentence:                                                                                                                                                                                                                                                    \\
          & \textcolor[rgb]{0.502,0.502,0.502}{L2} & ~When\textcolor{myblue}{ I will give a presentation at college,} I think \textcolor{myblue}{I will win the appreciation of my teachers and colleagues }since\textcolor{myorange}{ they clap their hands happily.~}                                                      \\
          & \textcolor[rgb]{0.502,0.502,0.502}{L3} & When \textcolor{myblue}{my boss takes me to a nice restaurant}, I think \textcolor{myblue}{he likes me and respects my opinion}\textcolor{myorange}{ }since\textcolor{myorange}{ I have gone out with him before and he has always seemed to enjoy my company.~}  \\
          & \textcolor[rgb]{0.502,0.502,0.502}{L4} & When \textcolor{myblue}{I will have a meeting with my boss,} I think\textcolor{myblue}{ it is an opportunity to show myself }since\textcolor{myorange}{ I have great knowledge about my work.~}                                                                         \\
          & \textcolor[rgb]{0.502,0.502,0.502}{L5} & When \textcolor{myblue}{\uline{Situation}}, I think \textcolor{myblue}{\uline{Thought}} since~                                                                                                                                                                                \\ 
\bottomrule
\toprule
\textbf{Type}                                                                                                     &                                        & \textbf{Prompt inputs for collecting~\textit{Actions}}                                                                                                                                                                                                                    \\ 
\midrule
\multirow{5}{*}{\begin{tabular}[c]{@{}c@{}} (a) \\ Action \\in the negative \\cognitive chain~\end{tabular}}       & \textcolor[rgb]{0.502,0.502,0.502}{L1}  & Complete the sentence:                                                                                                                                                                                                                                                    \\
          & \textcolor[rgb]{0.502,0.502,0.502}{L2}  & When \textcolor{myblue}{I will give a presentation at college} , I think \textcolor{myblue}{I will freeze on stage}, so \textcolor{myorange}{I rehearse in front of the mirror .~}                                                                                      \\
          & \textcolor[rgb]{0.502,0.502,0.502}{L3}  & When\textcolor{myblue}{ I was waiting in the waiting room for my last job interview}, I thought \textcolor{myblue}{I would not be selected}, so \textcolor{myorange}{I took a deep breath repeatedly.~}                                                                 \\
          & \textcolor[rgb]{0.502,0.502,0.502}{L4}  & When \textcolor{myblue}{my college entrance exam is tomorrow}, I think \textcolor{myblue}{I will fail the exam}, so \textcolor{myorange}{I scan through some old papers to review.~}                                                                                    \\
          & \textcolor[rgb]{0.502,0.502,0.502}{L5}  & When \textcolor{myblue}{\uline{Situation}}, I think \textcolor{myblue}{\uline{Thought}}, so~                                                                                                                                                                                  \\ 
\midrule
\multirow{5}{*}{\begin{tabular}[c]{@{}c@{}}(b) \\Action \\in the positive \\cognitive chain~\end{tabular}} & \textcolor[rgb]{0.502,0.502,0.502}{L1}  & ~Complete the
sentence:                                                                                                                                                                                                                                                 \\
          & \textcolor[rgb]{0.502,0.502,0.502}{L2}  & When \textcolor{myblue}{I will give a presentation at college,} I think\textcolor{myblue}{ I am going to ace this presentation}, so \textcolor{myorange}{I take good sleep at night.}                                                                                   \\
          & \textcolor[rgb]{0.502,0.502,0.502}{L3}  & When \textcolor{myblue}{I was waiting in the waiting room for my last job interview}, I thought\textcolor{myblue}{ I would land my dream job}, so \textcolor{myorange}{I talked to other candidates confidently.}                                                       \\
          & \textcolor[rgb]{0.502,0.502,0.502}{L4}  & When \textcolor{myblue}{my boss took me to a nice restaurant}, I thought \textcolor{myblue}{it was a good sign of a promotion}, so \textcolor{myorange}{I talked to him about my career plan.}                                                                          \\
          & \textcolor[rgb]{0.502,0.502,0.502}{L5}  & When \textcolor{myblue}{\uline{Situation}}, I think \uline{\textcolor{myblue}{Thought}}, so~                                                                                                                                                                                  \\ 

\bottomrule
\end{tabularx}}

\label{tab-promptforgentask}
\end{table*}

\clearpage

\begin{table*}[htbp]
\caption{Prompts for collecting \textbf{Emotion} labels in negative and positive cognitive chains. Each prompt input consists of 20 lines, and every 5 lines is a case.}
\centering
\renewcommand\arraystretch{1.5}
\scalebox{0.68}{
\begin{tabularx}{1.45\textwidth}{
p{.25\textwidth} <{\centering}
p{.03\textwidth} <{\raggedleft}
p{1.2\textwidth} <{\raggedright}}
\toprule

\textbf{Type} 
& & \textbf{\textbf{Prompt inputs for collecting~\textit{Emotions}}}                                                                                                                                                                                                          \\ 
\midrule
\multirow{20}{*}{\begin{tabular}[c]{@{}c@{}}(a) \\Emotion generation \\in the the negative chain~\end{tabular}}           & \textcolor[rgb]{0.502,0.502,0.502}{L1}  & Situation : \textcolor{myblue}{I arranged an interview for my friend with the CEO.}                                                                                                                                                                                         \\
          & \textcolor[rgb]{0.502,0.502,0.502}{L2}  & Thought :\textcolor{myblue}{ It will go badly.}                                                                                                                                                                                                                             \\
          & \textcolor[rgb]{0.502,0.502,0.502}{L3}  & Question : Choose one word to describe my feeling
about the Situation when I have the Thought.                                                                                                                                                                          \\
          & \textcolor[rgb]{0.502,0.502,0.502}{L4}  & Choice : Angry, Fearful, Sad                                                                                                                                                                                                                                              \\
          & \textcolor[rgb]{0.502,0.502,0.502}{L5}  & Answer : \textcolor{myorange}{Fearful}                                                                                                                                                                                                                              \\
          & \textcolor[rgb]{0.502,0.502,0.502}{L6}  & Situation : \textcolor{myblue}{The boss asserted his authority by yelling at the employees.}                                                                                                                                                                                \\
          & \textcolor[rgb]{0.502,0.502,0.502}{L7}  & Thought : \textcolor{myblue}{He's a terrible person.}                                                                                                                                                                                                                       \\
          & \textcolor[rgb]{0.502,0.502,0.502}{L8}  & Question : Choose one word to describe my feeling
about the Situation when I have the Thought.                                                                                                                                                                          \\
          & \textcolor[rgb]{0.502,0.502,0.502}{L9}  & Choice : Angry, Fearful, Sad                                                                                                                                                                                                                                              \\
          & \textcolor[rgb]{0.502,0.502,0.502}{L10} & Answer : \textcolor{myorange}{Angry}                                                                                                                                                                                                                                \\
          & \textcolor[rgb]{0.502,0.502,0.502}{L11} & Situation : \textcolor{myblue}{I asked my daughter to help me with the dishes.\textbackslash{}n}                                                                                                                                                                            \\
          & \textcolor[rgb]{0.502,0.502,0.502}{L12} & Thought :~\textcolor{myblue}{ she will refuse.}                                                                                                                                                                                                                             \\
          & \textcolor[rgb]{0.502,0.502,0.502}{L13} & Question : Choose one word to describe my feeling
about the Situation when I have the Thought.                                                                                                                                                                          \\
          & \textcolor[rgb]{0.502,0.502,0.502}{L14} & Choice : Angry, Fearful, Sad                                                                                                                                                                                                                                              \\
          & \textcolor[rgb]{0.502,0.502,0.502}{L15} & Answer : \textcolor{myorange}{Sad}                                                                                                                                                                                                                                  \\
          & \textcolor[rgb]{0.502,0.502,0.502}{L16} & Situation :~ \textcolor{myblue}{\uline{Situation Extended from Atomic Event}}                                                                                                                                                                                               \\
          & \textcolor[rgb]{0.502,0.502,0.502}{L17} & Thought : \textcolor{myblue}{\uline{Thought automatically generated and manually evaluated.}}                                                                                                                                                                               \\
          & \textcolor[rgb]{0.502,0.502,0.502}{L18} & Question : Choose one word to describe my feeling
about the Situation when I have the Thought.                                                                                                                                                                          \\
          & \textcolor[rgb]{0.502,0.502,0.502}{L19} & Choice : Angry, Fearful, Sad                                                                                                                                                                                                                                              \\
          & \textcolor[rgb]{0.502,0.502,0.502}{L20} & Answer:                                                                                                                                                                                                                                                                   \\ 
\midrule
\multirow{20}{*}{\begin{tabular}[c]{@{}c@{}}(b) \\Emotion generation\\in the the positive chain~\end{tabular}}            & \textcolor[rgb]{0.502,0.502,0.502}{L1}  & ~Situation :\textcolor{myblue}{ I arranged an interview for my friend with the CEO.}                                                                                                                                                                                        \\
          & \textcolor[rgb]{0.502,0.502,0.502}{L2}  & ~Thought :\textcolor{myblue}{ I will help my friend's career.}                                                                                                                                                                                                              \\
          & \textcolor[rgb]{0.502,0.502,0.502}{L3}  & ~Question :
Choose one word to describe my feeling about the Thought when I am in the
Situation.                                                                                                                                                                      \\
          & \textcolor[rgb]{0.502,0.502,0.502}{L4}  & ~Choice :
Joyful, Love, Surprised                                                                                                                                                                                                                                       \\
          & \textcolor[rgb]{0.502,0.502,0.502}{L5}  & ~Answer :~ \textcolor{myorange}{Joyful}                                                                                                                                                                                                                             \\
          & \textcolor[rgb]{0.502,0.502,0.502}{L6}  & ~Situation : \textcolor{myblue}{I also took out the new girl on a date.~}                                                                                                                                                                                                   \\
          & \textcolor[rgb]{0.502,0.502,0.502}{L7}  & ~Thought : \textcolor{myblue}{She will call me her boyfriend.}                                                                                                                                                                                                              \\
          & \textcolor[rgb]{0.502,0.502,0.502}{L8}  & ~Question :
Choose one word to describe my feeling about the Situation when I have the
Thought.                                                                                                                                                                       \\
          & \textcolor[rgb]{0.502,0.502,0.502}{L9}  & ~Choice :
Joyful, Love, Surprised                                                                                                                                                                                                                                       \\
          & \textcolor[rgb]{0.502,0.502,0.502}{L10} & ~Answer : \textcolor{myorange}{Love}                                                                                                                                                                                                                                \\
          & \textcolor[rgb]{0.502,0.502,0.502}{L11} & ~Situation : \textcolor{myblue}{I broke my leg skiing.}                                                                                                                                                                                                                     \\
          & \textcolor[rgb]{0.502,0.502,0.502}{L12} & ~Thought : \textcolor{myblue}{I will get a snowboard for my birthday.}                                                                                                                                                                                                      \\
          & \textcolor[rgb]{0.502,0.502,0.502}{L13} & ~Question :
Choose one word to describe my feeling about the Thought when I am in the
Situation.                                                                                                                                                                      \\
          & \textcolor[rgb]{0.502,0.502,0.502}{L14} & ~Choice :
Joyful, Love, Surprised                                                                                                                                                                                                                                       \\
          & \textcolor[rgb]{0.502,0.502,0.502}{L15} & ~Answer : \textcolor{myorange}{Surprised}                                                                                                                                                                                                                           \\
          & \textcolor[rgb]{0.502,0.502,0.502}{L16} & ~Situation:~ \textcolor{myblue}{\uline{Situation Extended from Atomic Event}}                                                                                                                                                                                                       \\
          & \textcolor[rgb]{0.502,0.502,0.502}{L17} & ~Thought: \textcolor{myblue}{\uline{Thought automatically generated and manually evaluated.}}                                                                                                                                                                                       \\
          & \textcolor[rgb]{0.502,0.502,0.502}{L18} & ~Question:
Choose one word to describe my feeling about the Thought when I am in the
Situation.                                                                                                                                                                       \\
          & \textcolor[rgb]{0.502,0.502,0.502}{L19} & ~Choice: Joyful,
Love, Surprised                                                                                                                                                                                                                                        \\
          & \textcolor[rgb]{0.502,0.502,0.502}{L20} & Answer:  \\
\bottomrule
\end{tabularx}}
\label{tab-promptforemotask}
\end{table*}

\clearpage

%% file: acl_latex.bbl
\begin{thebibliography}{39}
\expandafter\ifx\csname natexlab\endcsname\relax\def\natexlab#1{#1}\fi

\bibitem[{Achiam et~al.(2023)Achiam, Adler, Agarwal, Ahmad, Akkaya, Aleman, Almeida, Altenschmidt, Altman, Anadkat et~al.}]{achiam2023gpt}
Josh Achiam, Steven Adler, Sandhini Agarwal, Lama Ahmad, Ilge Akkaya, Florencia~Leoni Aleman, Diogo Almeida, Janko Altenschmidt, Sam Altman, Shyamal Anadkat, et~al. 2023.
\newblock Gpt-4 technical report.
\newblock \emph{arXiv preprint arXiv:2303.08774}.

\bibitem[{Apperly(2010)}]{apperly2010mindreaders}
Ian Apperly. 2010.
\newblock \emph{Mindreaders: the cognitive basis of" theory of mind"}.

\bibitem[{Baldwin(1992)}]{baldwin1992relational}
Mark~W Baldwin. 1992.
\newblock Relational schemas and the processing of social information.
\newblock \emph{Psychological bulletin}, page 461.

\bibitem[{Bosselut et~al.(2019)Bosselut, Rashkin, Sap, Malaviya, Celikyilmaz, and Choi}]{BosselutRSMCC19}
Antoine Bosselut, Hannah Rashkin, Maarten Sap, Chaitanya Malaviya, Asli Celikyilmaz, and Yejin Choi. 2019.
\newblock \href {https://doi.org/10.18653/v1/p19-1470} {{COMET:} commonsense transformers for automatic knowledge graph construction}.
\newblock In \emph{ACL}, pages 4762--4779.

\bibitem[{Call and Tomasello(2011)}]{call2011tom30year}
Josep Call and Michael Tomasello. 2011.
\newblock Does the chimpanzee have a theory of mind? 30 years later.
\newblock \emph{Human Nature and Self Design}, pages 83--96.

\bibitem[{Choi(2022)}]{choi2022curious}
Yejin Choi. 2022.
\newblock The curious case of commonsense intelligence.
\newblock \emph{Daedalus}, pages 139--155.

\bibitem[{Dhelim et~al.(2021)Dhelim, Ning, Farha, Chen, Atzori, and Daneshmand}]{Dhelim2021IoTEnabledSR}
Sahraoui Dhelim, Huansheng Ning, Fadi Farha, Liming~Luke Chen, Luigi Atzori, and Mahmoud Daneshmand. 2021.
\newblock Iot-enabled social relationships meet artificial social intelligence.
\newblock \emph{IEEE Internet of Things Journal}, pages 17817--17828.

\bibitem[{Harris et~al.(1989)Harris, Johnson, Hutton, Andrews, and Cooke}]{harris1989young}
Paul~L Harris, Carl~N Johnson, Deborah Hutton, Giles Andrews, and Tim Cooke. 1989.
\newblock Young children's theory of mind and emotion.
\newblock \emph{Cognition \& Emotion}, pages 379--400.

\bibitem[{Hu et~al.(2021)Hu, Shen, Wallis, Allen-Zhu, Li, Wang, Wang, and Chen}]{hu2021lora}
Edward~J Hu, Yelong Shen, Phillip Wallis, Zeyuan Allen-Zhu, Yuanzhi Li, Shean Wang, Lu~Wang, and Weizhu Chen. 2021.
\newblock Lora: Low-rank adaptation of large language models.
\newblock \emph{arXiv preprint arXiv:2106.09685}.

\bibitem[{Hwang et~al.(2021)Hwang, Bhagavatula, Le~Bras, Da, Sakaguchi, Bosselut, and Choi}]{hwang2021symbolic}
Jena~D Hwang, Chandra Bhagavatula, Ronan Le~Bras, Jeff Da, Keisuke Sakaguchi, Antoine Bosselut, and Yejin Choi. 2021.
\newblock On symbolic and neural commonsense knowledge graphs.

\bibitem[{Kim et~al.(2023)Kim, Hessel, Jiang, West, Lu, Yu, Zhou, Bras, Alikhani, Kim, Sap, and Choi}]{kim2023soda}
Hyunwoo Kim, Jack Hessel, Liwei Jiang, Peter West, Ximing Lu, Youngjae Yu, Pei Zhou, Ronan~Le Bras, Malihe Alikhani, Gunhee Kim, Maarten Sap, and Yejin Choi. 2023.
\newblock Soda: Million-scale dialogue distillation with social commonsense contextualization.
\newblock In \emph{EMNLP}.

\bibitem[{Langley et~al.(2022)Langley, Cirstea, Cuzzolin, and Sahakian}]{Langley2022TheoryOM}
Christelle Langley, Bogdan-Ionut Cirstea, Fabio Cuzzolin, and Barbara~Jacquelyn Sahakian. 2022.
\newblock Theory of mind and preference learning at the interface of cognitive science, neuroscience, and ai: A review.
\newblock \emph{Frontiers in Artificial Intelligence}.

\bibitem[{Lavie and Denkowski(2009)}]{LavieD09}
Alon Lavie and Michael~J. Denkowski. 2009.
\newblock \href {https://doi.org/10.1007/s10590-009-9059-4} {The meteor metric for automatic evaluation of machine translation}.
\newblock \emph{Mach. Transl.}, 23(2-3):105--115.

\bibitem[{Le et~al.(2019)Le, Boureau, and Nickel}]{Le2019RevisitingTE}
Matt Le, Y-Lan Boureau, and Maximilian Nickel. 2019.
\newblock Revisiting the evaluation of theory of mind through question answering.
\newblock In \emph{EMNLP}.

\bibitem[{Leslie et~al.(2004)Leslie, Friedman, and German}]{leslie2004core}
Alan~M Leslie, Ori Friedman, and Tim~P German. 2004.
\newblock Core mechanisms in ‘theory of mind’.
\newblock \emph{Trends in cognitive sciences}, pages 528--533.

\bibitem[{Li et~al.(2017)Li, Su, Shen, Li, Cao, and Niu}]{LiSSLCN17}
Yanran Li, Hui Su, Xiaoyu Shen, Wenjie Li, Ziqiang Cao, and Shuzi Niu. 2017.
\newblock \href {https://aclanthology.org/I17-1099/} {Dailydialog: {A} manually labelled multi-turn dialogue dataset}.
\newblock In \emph{IJCNLP}, pages 986--995.

\bibitem[{Lin(2004)}]{lin2004rouge}
Chin-Yew Lin. 2004.
\newblock Rouge: A package for automatic evaluation of summaries.
\newblock In \emph{Text summarization branches out}, pages 74--81.

\bibitem[{Liu et~al.(2021)Liu, Zheng, Demasi, Sabour, Li, Yu, Jiang, and Huang}]{DBLP:conf/acl/LiuZDSLYJH20}
Siyang Liu, Chujie Zheng, Orianna Demasi, Sahand Sabour, Yu~Li, Zhou Yu, Yong Jiang, and Minlie Huang. 2021.
\newblock \href {https://doi.org/10.18653/v1/2021.acl-long.269} {Towards emotional support dialog systems}.
\newblock In \emph{ACL/IJCNLP 2021}, pages 3469--3483.

\bibitem[{Loshchilov and Hutter(2019)}]{LoshchilovH19}
Ilya Loshchilov and Frank Hutter. 2019.
\newblock \href {https://openreview.net/forum?id=Bkg6RiCqY7} {Decoupled weight decay regularization}.
\newblock In \emph{ICLR}. OpenReview.net.

\bibitem[{Mehl et~al.(2020)Mehl, Hesse, Schmidt, Landsberg, Soll, Bechdolf, Herrlich, Kircher, Klingberg, M{\"u}ller et~al.}]{mehl2020theory}
Stephanie Mehl, Klaus Hesse, Anna-Christine Schmidt, Martin~W Landsberg, Daniel Soll, Andreas Bechdolf, Jutta Herrlich, Tilo Kircher, Stefan Klingberg, Bernhard~W M{\"u}ller, et~al. 2020.
\newblock Theory of mind, emotion recognition, delusions and the quality of the therapeutic relationship in patients with psychosis--a secondary analysis of a randomized-controlled therapy trial.
\newblock \emph{BMC psychiatry}, pages 1--13.

\bibitem[{Meinhardt-Injac et~al.(2018)Meinhardt-Injac, Daum, Meinhardt, and Persike}]{meinhardt2018two}
Bozana Meinhardt-Injac, Moritz~M Daum, G{\"u}nter Meinhardt, and Malte Persike. 2018.
\newblock The two-systems account of theory of mind: Testing the links to social-perceptual and cognitive abilities.
\newblock \emph{Frontiers in human neuroscience}, page~25.

\bibitem[{Mostafazadeh et~al.(2020)Mostafazadeh, Kalyanpur, Moon, Buchanan, Berkowitz, Biran, and Chu-Carroll}]{mostafazadeh-etal-2020-glucose}
Nasrin Mostafazadeh, Aditya Kalyanpur, Lori Moon, David Buchanan, Lauren Berkowitz, Or~Biran, and Jennifer Chu-Carroll. 2020.
\newblock \href {https://doi.org/10.18653/v1/2020.emnlp-main.370} {{GLUCOSE}: {G}enera{L}ized and {CO}ntextualized story explanations}.
\newblock In \emph{EMNLP}, pages 4569--4586.

\bibitem[{Ouyang et~al.(2022)Ouyang, Wu, Jiang, Almeida, Wainwright, Mishkin, Zhang, Agarwal, Slama, Ray et~al.}]{ouyang2022training}
Long Ouyang, Jeffrey Wu, Xu~Jiang, Diogo Almeida, Carroll Wainwright, Pamela Mishkin, Chong Zhang, Sandhini Agarwal, Katarina Slama, Alex Ray, et~al. 2022.
\newblock Training language models to follow instructions with human feedback.
\newblock \emph{Advances in Neural Information Processing Systems}.

\bibitem[{Papineni et~al.(2002)Papineni, Roukos, Ward, and Zhu}]{Papineni2002BleuAM}
Kishore Papineni, Salim Roukos, Todd Ward, and Wei-Jing Zhu. 2002.
\newblock Bleu: a method for automatic evaluation of machine translation.
\newblock In \emph{ACL}.

\bibitem[{Phillip et~al.(1987)Phillip, Judith, Donald, and O'connor}]{shaver1987emotion}
Shaver Phillip, Schwartz Judith, Kirson Donald, and Cary O'connor. 1987.
\newblock Emotion knowledge: further exploration of a prototype approach.
\newblock \emph{Journal of personality and social psychology}, page 1061.

\bibitem[{Premack and Woodruff(1978)}]{premack1978tom}
David Premack and Guy Woodruff. 1978.
\newblock Does the chimpanzee have a theory of mind?
\newblock \emph{Behavioral and brain sciences}, pages 515--526.

\bibitem[{Rabinowitz et~al.(2018)Rabinowitz, Perbet, Song, Zhang, Eslami, and Botvinick}]{RabinowitzPSZEB18}
Neil~C. Rabinowitz, Frank Perbet, H.~Francis Song, Chiyuan Zhang, S.~M.~Ali Eslami, and Matthew~M. Botvinick. 2018.
\newblock \href {http://proceedings.mlr.press/v80/rabinowitz18a.html} {Machine theory of mind}.
\newblock In \emph{ICML}, pages 4215--4224.

\bibitem[{Roller et~al.(2021)Roller, Dinan, Goyal, Ju, Williamson, Liu, Xu, Ott, Smith, Boureau, and Weston}]{DBLP:conf/eacl/RollerDGJWLXOSB21}
Stephen Roller, Emily Dinan, Naman Goyal, Da~Ju, Mary Williamson, Yinhan Liu, Jing Xu, Myle Ott, Eric~Michael Smith, Y{-}Lan Boureau, and Jason Weston. 2021.
\newblock \href {https://doi.org/10.18653/v1/2021.eacl-main.24} {Recipes for building an open-domain chatbot}.
\newblock In \emph{EACL 2021}, pages 300--325.

\bibitem[{Sap et~al.(2022)Sap, Bras, Fried, and Choi}]{Sap2022NeuralTO}
Maarten Sap, Ronan~Le Bras, Daniel Fried, and Yejin Choi. 2022.
\newblock Neural theory-of-mind? on the limits of social intelligence in large lms.
\newblock \emph{ArXiv}.

\bibitem[{Sap et~al.(2019)Sap, Le~Bras, Allaway, Bhagavatula, Lourie, Rashkin, Roof, Smith, and Choi}]{Sap2019ATOMICAA}
Maarten Sap, Ronan Le~Bras, Emily Allaway, Chandra Bhagavatula, Nicholas Lourie, Hannah Rashkin, Brendan Roof, Noah~A Smith, and Yejin Choi. 2019.
\newblock Atomic: An atlas of machine commonsense for if-then reasoning.
\newblock In \emph{AAAI}, pages 3027--3035.

\bibitem[{Shapira et~al.(2023)Shapira, Levy, Alavi, Zhou, Choi, Goldberg, Sap, and Shwartz}]{shapira2023clever}
Natalie Shapira, Mosh Levy, Seyed~Hossein Alavi, Xuhui Zhou, Yejin Choi, Yoav Goldberg, Maarten Sap, and Vered Shwartz. 2023.
\newblock Clever hans or neural theory of mind? stress testing social reasoning in large language models.
\newblock \emph{arXiv preprint arXiv:2305.14763}.

\bibitem[{Speer et~al.(2016)Speer, Chin, and Havasi}]{Speer2016ConceptNet5A}
Robyn Speer, Joshua Chin, and Catherine Havasi. 2016.
\newblock Conceptnet 5.5: An open multilingual graph of general knowledge.
\newblock \emph{ArXiv}.

\bibitem[{Touvron et~al.(2023)Touvron, Martin, Stone, Albert, Almahairi, Babaei, Bashlykov, Batra, Bhargava, Bhosale et~al.}]{touvron2023llama}
Hugo Touvron, Louis Martin, Kevin Stone, Peter Albert, Amjad Almahairi, Yasmine Babaei, Nikolay Bashlykov, Soumya Batra, Prajjwal Bhargava, Shruti Bhosale, et~al. 2023.
\newblock Llama 2: Open foundation and fine-tuned chat models.
\newblock \emph{arXiv preprint arXiv:2307.09288}.

\bibitem[{Wang et~al.(2023)Wang, Li, Yin, Wu, and Liu}]{wang2023emotional}
Xuena Wang, Xueting Li, Zi~Yin, Yue Wu, and Jia Liu. 2023.
\newblock Emotional intelligence of large language models.
\newblock \emph{Journal of Pacific Rim Psychology}.

\bibitem[{West et~al.(2021)West, Bhagavatula, Hessel, Hwang, Jiang, Bras, Lu, Welleck, and Choi}]{west2021symbolic}
Peter West, Chandra Bhagavatula, Jack Hessel, Jena~D Hwang, Liwei Jiang, Ronan~Le Bras, Ximing Lu, Sean Welleck, and Yejin Choi. 2021.
\newblock Symbolic knowledge distillation: from general language models to commonsense models.
\newblock \emph{arXiv preprint arXiv:2110.07178}.

\bibitem[{West et~al.(2023)West, Bras, Sorensen, Lin, Jiang, Lu, Chandu, Hessel, Baheti, Bhagavatula et~al.}]{west2023novacomet}
Peter West, Ronan~Le Bras, Taylor Sorensen, Bill~Yuchen Lin, Liwei Jiang, Ximing Lu, Khyathi Chandu, Jack Hessel, Ashutosh Baheti, Chandra Bhagavatula, et~al. 2023.
\newblock Novacomet: Open commonsense foundation models with symbolic knowledge distillation.
\newblock \emph{arXiv preprint arXiv:2312.05979}.

\bibitem[{Westbrook et~al.(2011)Westbrook, Kennerley, and Kirk}]{Westbrook2007AnIT}
David Westbrook, Helen Kennerley, and Joan Kirk. 2011.
\newblock \emph{An introduction to cognitive behaviour therapy: Skills and applications}.

\bibitem[{Zalla and Korman(2018)}]{zalla2018prior}
Tiziana Zalla and Joanna Korman. 2018.
\newblock Prior knowledge, episodic control and theory of mind in autism: Toward an integrative account of social cognition.
\newblock \emph{Frontiers in psychology}.

\bibitem[{Zhang et~al.(2019)Zhang, Kishore, Wu, Weinberger, and Artzi}]{zhang2019bertscore}
Tianyi Zhang, Varsha Kishore, Felix Wu, Kilian~Q Weinberger, and Yoav Artzi. 2019.
\newblock Bertscore: Evaluating text generation with bert.
\newblock \emph{arXiv preprint arXiv:1904.09675}.

\end{thebibliography}
